\documentclass[10pt,twocolumn,letterpaper]{article}

\usepackage[pagenumbers]{cvpr} 

\usepackage[dvipsnames]{xcolor}

\usepackage{booktabs}
\usepackage{adjustbox}
\newcolumntype{R}[2]{%
    >{\adjustbox{angle=#1,lap=\width-(#2)}\bgroup}%
    l%
    <{\egroup}%
}

\definecolor{cvprblue}{rgb}{0.21,0.49,0.74}
\usepackage[pagebackref,breaklinks,colorlinks,citecolor=cvprblue]{hyperref}

\def\paperID{21} 
\def\confName{CVPR}
\def\confYear{2024}

\title{COOD: Combined out-of-distribution detection using multiple measures for anomaly \& novel class detection in large-scale hierarchical classification}

\author{Laurens E. Hogeweg\\
Intel Benelux BV / Naturalis Biodiversity Center\\
HTC 83, 5656 AE Eindhoven, The Netherlands / Darwinweg 2, 2333 CR Leiden, The Netherlands\\
{\tt\small laurens.hogeweg@intel.com / laurens.hogeweg@naturalis.nl}
\and
Rajesh Gangireddy\\
Intel Benelux BV\\
{\tt\small rajesh.gangireddy@intel.com}
\and
Django Brunink\\
Naturalis Biodiversity Center\\
{\tt\small django.brunink@naturalis.nl}
\and
Vincent J. Kalkman\\
Naturalis Biodiversity Center\\
{\tt\small vincent.kalkman@naturalis.nl}
\and
Ludo Cornelissen\\
Intel Benelux BV\\
{\tt\small ludo.cornelissen@intel.com}
\and
Jacob W. Kamminga\\
University of Twente\\
Drienerlolaan 5, 7522 NB Enschede, The Netherlands\\
{\tt\small j.w.kamminga@utwente.nl}
}

\begin{document}
\maketitle
\begin{abstract}
High-performing out-of-distribution (OOD) detection, both anomaly and novel class, is an important prerequisite for the practical use of classification models. In this paper, we focus on the species recognition task in images concerned with large databases, a large number of fine-grained hierarchical classes, severe class imbalance, and varying image quality. We propose a framework for combining individual OOD measures into one combined OOD (\textit{COOD}) measure using a supervised model. The individual measures are several existing state-of-the-art measures and several novel OOD measures developed with novel class detection and hierarchical class structure in mind. COOD was extensively evaluated on three large-scale (500k+ images) biodiversity datasets in the context of anomaly and novel class detection. We show that COOD outperforms individual, including state-of-the-art, OOD measures by a large margin in terms of TPR@1\% FPR in the majority of experiments, e.g., improving detecting ImageNet images (OOD) from 54.3\% to 85.4\% for the iNaturalist 2018 dataset. SHAP (feature contribution) analysis shows that different individual OOD measures are essential for various tasks, indicating that multiple OOD measures and combinations are needed to generalize. Additionally, we show that explicitly considering ID images that are incorrectly classified for the original (species) recognition task is important for constructing high-performing OOD detection methods and for practical applicability. The framework can easily be extended or adapted to other tasks and media modalities.
\end{abstract}    
\section{Introduction}
\label{sec:intro}

In recent years, the application of deep learning for image classification tasks has yielded remarkable accuracies across various domains. Many of these high-accuracy classification models might become unreliable due to the lack of knowledge of open and changing environments when used in real-world applications~\cite{AmodeiOSCSM16, nguyen2015deep}. For instance, in species recognition~\cite{anubha2019study}, when an image of a new species that the classification model has not been trained to classify is encountered, it is better to reject it as “unknown” or out-of-distribution (OOD) than to (mis)classify it into one of the known classes~\cite{nguyen2015deep}. OOD detection is even more critical in fields such as medicine~\cite{zimmerer9762702} and autonomous driving \cite{di9525313}. In interactive applications, such as mobile apps for species recognition, the OOD system can help the user take better pictures and avoid the submission of unusable input such as selfies, landscapes, etc.  In (semi-)~automated computer vision applications, rejecting OOD inputs while rejecting a minimum amount of in-distribution (ID) samples is crucial.   

As OOD detection is increasingly used as a prerequisite for open-world computer vision applications, there has been a growing interest in this domain in recent years \cite{abdar2021review, ruff2021unifying, liu2021towards}.  Several new OOD measures (for instance \cite{sun2022out, ndiour2020out, Liu2020EnergybasedOD}) have been explored and have led to state-of-the-art OOD detection methods. However, most of the OOD detection methods are benchmarked on a limited set of small datasets with OOD datasets from an entirely different domain \cite{abs-2109-05554, abs-1910-14034}. More importantly, different OOD measures might have properties which make them perform  well on selected OOD datasets and perform less well in other OOD scenarios \cite{abs-2109-05554}. 

While developing individual state-of-the-art measures is worthwhile, in many cases, individual methods will have particular strengths and weaknesses. Therefore, the combination of several well-performing methods could very well outperform the individual methods as a weakness of one method is canceled by the strength of another. Many examples from machine learning literature show that a combination of methods outperforms individual ones \cite{mohandes2018classifiers, ganaie2022ensemble}. 

Here, we introduce the Combined OOD (\textit{\textbf{COOD}}) measure framework: a learned combination of existing and novel individual OOD measures which combines the strengths of OOD measures to overcome the limitations of others. We show that COOD significantly outperforms the best individual OOD measures in different OOD scenarios, which test various levels of OOD detection difficulty – near, intermediate (mid), and far OOD.  We also introduce several novel OOD measures that exploit the hierarchical class structure and the discrepancy between linear and kNN predictions. Our method is supervised and requires external datasets to train. Although a slight disadvantage compared to methods that only use ID data, the use of OOD data has proven popular \cite{hendrycks2018deep} and has advantages such as making operating point calibration easier.
 
In terms of testing our framework, we focus on OOD detection tasks for the biodiversity domain due to the many challenges they pose. Specifically, we focus on large-scale datasets (500k+ images) where ID classes have a hierarchical structure. Biodiversity datasets have challenging properties such as high-class imbalance (long-tailed), where certain species (classes) are much more abundant or better represented in the data than most. Encountering novel classes is common in biodiversity classification applications due to different geographic distributions, incomplete databases, and other factors. Due to the species' fine-grained (visually similar) nature, expert knowledge is required to label new classes accurately. The limited data can make it harder for neural networks to generalize well and could classify OOD data into one of the long-tailed classes with high confidence. 

To summarise, we make the following contributions: (1) COOD – a novel framework that combines existing and novel OOD measures. Extensive evaluation shows improved OOD detection on three biodiversity datasets. (2) Several novel OOD measures focused on hierarchical labels and novel class detection (3) show that explicitly defining how to deal with ID but incorrect predictions is important for consistent analysis and application in practice. These improvements allow a more robust implementation of classification models in practical settings: rejecting unusable inputs and finding novel classes, such as rare species or rare diseases, more reliably.


\section{Data}
\label{sec:data}

\label{sec:datasets}
Two sources of data were used in this paper: (1) Multi-Source-Model (MSM): a large-scale dataset from field observations of organisms in Europe (2) the iNaturalist 2018 large scale fine-grained dataset \cite{inat_dataset}. The MSM model aims to identify field observations, typically from mobile phones, of organisms from Europe at large scale \cite{schermer2018supporting}. It consists of a top-level model for broad classification and sub-models for fine-grained classification of species. Additionally, specialised models were trained per source of data (Norway, Sweden, Denmark, UK, rest of Europe). This structure of the models and the scale of the dataset (33M images in total) allows us to do large-scale novel class detection. Existing trained models to classify taxa (species) in the datasets were used for the experiments. Three datasets were defined for the paper.


(1) The \textbf{MSM top-level} model categories field observations into 8 categories (plants, fungi, vertebrates, butterflies \& moths, flies, other insects, other arthropods, other invertebrates) and has a top-1 accuracy of 93.7\%. The dataset consists of 507,904 images. For the OOD dataset we use ImageNet \cite{ImageNet_dataset} where we exclude images tagged as “organism” (\textit{OOD-far: ImageNet-Non-Organism}; images outside the domain of biodiversity, 28,801 images). To determine the influence of domain overlap we used the cars from ImageNet (\textit{OOD-far: ImageNet-Cars}; 1,000 images) as a relatively easy OOD dataset.


(2) The \textbf{Norwegian vertebrates} (birds, mammals, reptiles, etc.) MSM sub-model classifies field observations into 972 taxa (biological classes) occurring in Norway and has a top-1 accuracy of 86.3\%. The dataset consists of 628,713 images. We use \textit{OOD-far: ImageNet-Non-Organism} for the OOD dataset. Two additional datasets were used for novel class detection: (1) 
 \textit{OOD-near: non-Norwegian vertebrates} (closely related in-domain classes; 1,123 images/novel classes) (2) \textit{OOD-mid: Norwegian non-vertebrates} (more distinct in-domain classes; 28,629 images/novel classes).


(3) \textbf{iNaturalist 2018} \cite{inat_dataset}  is a biodiversity dataset with 437,513 images for training and 24,426 images for validation. The dataset has 8,142 classes of fine-grained (visually similar) species spanning various taxonomic groups, including but not limited to plants, animals, fungi, and insects. \textit{OOD-far: ImageNet-Non-Organism} is used as the OOD dataset.

\section{Methods}
\label{sec:methods}

\subsection{OOD detection}
\label{sec:ood_methods}
Instead of aiming to develop a single state-of-the-art measure for performing OOD detection we present a framework which combines multiple state-of-the-art OOD measures - including several novel measures - into one Combined OOD  measure  (“COOD”).

For every image a feature vector is computed by global average pooling the output of the last convolutional block. From the feature vector the logits were computed by multiplying with the classification weight matrix $W$ and applying the bias $b$ (\Cref{tab:basic_math_def} for details). The linear probability vector was computed by applying \textit{SoftMax}. For all images the true label is known.

A kNN model forms the basis for many of the OOD measures. For every query point we calculate the $k=30$ nearest neighbors (NN). The inner product was used as distance measure and PCA with 256 components was applied as a pre-processing step. For the index we used Flat with an inverted file structure (IVF256). The implementation by FAISS was used \cite{douze2024faiss}. The neighbors are samples from the training set, and we have information about both the predictions and the true label. From the NN, we derive a kNN class probability vector by counting the true classes among the neighbors and normalising to 1. 

When using hierarchical classes, such as biological taxa, a measure can be defined of how conceptually different two classes are by computing a distance between them. This distance is defined as the weighted number of edges in the shortest path between two class nodes in the hierarchical tree (\Cref{fig:taxon_distance}). High class node (taxon) distances between kNN and linear predictions indicate that their results are completely different (e.g. one predicting a plant, the other a bird) while low (non-zero) values indicate that the two predictions almost agree (e.g. confusing two species of bird from the same taxonomical genus).
\subsection{OOD Measures}
Table \ref{tab:ood_measures} lists the 19 individual OOD measures that were used in the method. Some of them are existing methods, including state-of-the-art methods. Several of them are novel to our knowledge. A few other are components of other measures (e.g. \textit{Avg. distance among neighbors}) which might contribute to OOD detection.

\begin{table*}[t]
  \centering
  \begin{tabularx}{\linewidth}{>{\hsize=0.37\hsize}X>{\hsize=1.2\hsize}X>{\hsize=0.3\hsize}X}
  \midrule
    OOD Measure & Description & Source \\
  \toprule
    \textit{Avg. distance among NN}  & If the average distance among NN is high the query point lies in a low-density valley with NN scattered around it.  & Component from~\cite{zhang2009new}  \\
    \textit{Avg. distance to NN}  & If the average distance to the neighbors is high the query point lies far away from training data  & Component \cite{zhang2009new} \\
    \textit{Distance to 1st NN}  & The distance to the 1st neighbor is indicative how much a query feature deviates from the training set.   & \cite{bergman2020deep}  \\
    \textit{Distance to k-th NN}  & If the k-th neighbor is far away the query point is distinct from related images, similar but potentially less sensitive to noise as previous  & \cite{sun2022out}  \\
    \textit{LDOF} & Local distance outlier factor = \textit{Avg. distance to NN} / \textit{Avg. distance among NN} & \cite{zhang2009new}  \\
    \textit{Global FRE}  & Reconstruction error of the feature after applying a PCA model trained on all ID features & \cite{ndiour2020out}  \\
    \textit{Class FRE}  & Reconstruction error of the feature after applying a PCA model trained on ID features for the predicted label of the query image & \cite{ndiour2020out} \\
    \textit{Max(linear)}  & Maximum probability of the original linear prediction. In a calibrated model low probabilities indicate uncertainty.   & \cite{hendrycks2018baseline}  \\
    \textit{Max(knn)}  & Idem as previous but computed from the kNN probability vector  & \cite{yousefnezhad2021ensemble} \\
    \textit{Max(linear-T-scaled)}  & Probability computed using softmax with a temperature of 2.0 to reduce over-confidence and improve OOD detection  & \cite{liang2020enhancing}  \\
    \textit{Max(linear+kNN)}  & The maximum probability of the average of linear and kNN probability vectors, indicating agreement/disagreement (high/low values) between the linear and kNN predictions.   & Novel  \\
    \textit{TD(linear, kNN)}  & Conceptual distance between linear and kNN predicted labels computed using the taxon distance (\Cref{sec:ood_methods}).  & Novel  \\
    \textit{Entropy of NN’s true class}  & The variation among NN's true class is calculated using  entropy. & Component \cite{gangireddy2023}   \\
    \textit{EnWeDi(1st)}  & \textit{Distance to 1st neighbor} is weighted by  \textit{1 + Entropy of NN’s true class}  & \cite{gangireddy2023}  \\
    \textit{EnWeDi(average)}  & \textit{Average distance to
NN} is weighted by \textit{1 + Entropy of NN’s true class} & Novel, from \cite{gangireddy2023} \\
    \textit{Feature entropy} & For ID images feature values could be more concentrated relative to OOD, indicating the presence of class-specific image features. Measured by computing the entropy of the normalised feature vector.  & \cite{diwan2021feature}  \\
    \textit{Feature sum}  & In OOD features there might be an absence of feature responses compared to ID features, measured as the sum of the absolute feature values.  & Novel \\
    \textit{Feature magnitude}  & OOD samples might have very low or very high feature values. Measured by the length of the feature vector. & Component \cite{tian2021weakly} \\    
    \textit{Avg. true probability of NN}  & If many of the NN have low true probabilities for the true class, this implies that similar images as the query image are hard to classify correctly.  & Novel  \\
    \bottomrule
  \end{tabularx}
  \caption{Overview of individual OOD measures. NN = nearest neighbors. FRE = feature reconstruction error \cite{ndiour2020out}, PCA = principal component analysis. EnWeDi = Entropy Weighted Distance \cite{gangireddy2023}, TD = taxon distance (\Cref{sec:ood_methods}), mathematical definitions in \Cref{tab:ood_measures_app}}
  \label{tab:ood_measures}
\end{table*}

\subsection{COOD: Supervised combination of individual measures}
The different OOD measures are combined into one combined OOD score (\textit{COOD}) (0 = ID, 1 = OOD) using a RandomForest classifier. RandomForest is a popular method for tabular data with good properties in terms of overfitting resistance and limited sensitivity to class imbalance \cite{ho1995random}. Using a classifier allows to exploit (non-linear) relationships between OOD measures. The default setting of the scikit-learn v0.24.2 implementation was used.

\section{Experiments}

\subsection{Classification models}
A standard  neural network configuration was used where a backbone computes a feature vector which is mapped into prediction space using a dense classification layer - equivalent to a applying a linear model. MSM models were trained using an EfficientnetV2M \cite{tan2019efficientnet} architecture, with a cosine warmup strategy (startup phase of 2 epochs, a plateau of 4 epochs and a cosine phase of 30 epochs). Class balancing was used during training to improve classification of minority classes. For the iNaturalist dataset, the InceptionV3 \cite{inceptionv3_paper} model provided by the iNaturalist 2018 Competition \cite{inat2018_pretrained_model_github_repo} was used, which is reported to have a top one accuracy of 60.20\% on the validation set.

\subsection{Train/validation split}
To compute many of the OOD measures (e.g. kNN-based and FRE) a training set is needed. The training/validation split of the original classification task was used for this (90\%/10\% for MSM and for iNaturalist 2018 as published). All subsequent OOD measure computation and experiments are done on the original task's validation subset. This subset was split in training/validation (80\%/20\%)  again, resulting in 8\% of the original dataset used for training and 2\% for validation. Because no hyperparameter optimisation or early stopping was involved in the OOD experiments we report results on the OOD validation split directly. 

\subsection{Definition of reference}
\label{sec:ref_def}
For each OOD detection experiment the ID and OOD reference needs to be defined. The OOD datasets are external datasets which should be rejected by the OOD model. The ID dataset is further refined into several categories (\Cref{tab:id_cat_def}). These extra ID categories are used to show that the definition of positive and negative for both the OOD model training and model evaluation is important and relevant for practical applications. 

\begin{table}[]
    \centering
    \begin{tabularx}{\linewidth}{p{\dimexpr.25\linewidth-2\tabcolsep-1.3333\arrayrulewidth}p{\dimexpr.75\linewidth-2\tabcolsep-1.3333\arrayrulewidth}}
        \toprule
        ID-correct  & in-distribution (ID) image for which the original classifier’s prediction is correct \\
        \midrule
        ID-incorrect-high  & ID image for which the original classifier’s prediction is incorrect, \textit{Max(linear)} is \textgreater 80\% and the taxon distance (TD; \Cref{sec:ood_methods}) between correct and incorrect taxon is \textgreater 4. TD \textgreater 4 means correct and incorrect taxa are not closely related, corresponding to the group of highly confident (very) wrong predictions \cite{nguyen2015deep}\\
        \midrule
        ID-incorrect & the remainder of the ID images for which the original classifier’s prediction is incorrect: \textit{Max(linear)} \textless 80\% or TD \textless 4\\
        \bottomrule
    \end{tabularx}
    \caption{Definition of ID-categories}
    \label{tab:id_cat_def}
\end{table}

\subsection{Evaluation measures}
True positives (TP) are defined as OOD images being correctly detected/rejected as OOD after applying a threshold to the (C)OOD measure. False positives are defined as ID samples being incorrectly rejected as OOD. The two main evaluation measures are (1) \textbf{TPR@1\%FPR} = \% OOD detected @ 1\% ID rejected = \% OOD detected @ 99\% ID accepted (2) \textbf{AUROC} = Area under the ROC curve. We chose these definitions because we consider \% ID rejected (FPR) as the independent (control) variable in ROC analysis and as most important for practical applications.

\section{Results}
\subsection{Performance of individual OOD measures}
\begin{figure}[h]
    \centering
    \includegraphics[width=1.0\linewidth]{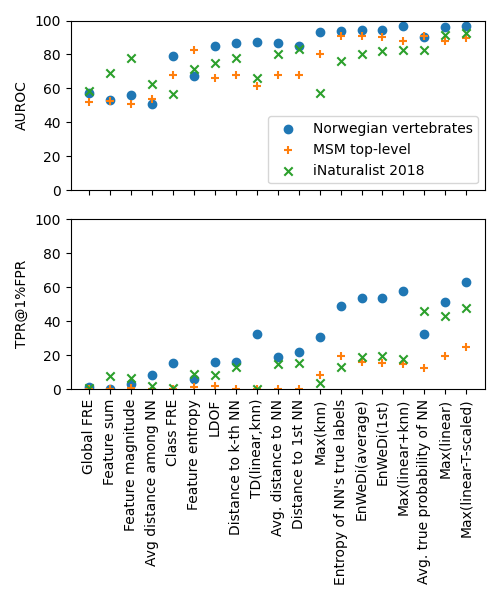}
    \caption{Performance of individual OOD measures for different datasets. The individual OOD measures are sorted by the average TPR@1\%FPR across datasets.}
    \label{fig:best_individual_plot}
\end{figure}

\Cref{fig:best_individual_plot} shows the performance of individual OOD measures for the three datasets used. The measures are ranked left to right by the average TPR@1\%FPR across datasets. The Norwegian vertebrates dataset has in general higher scores than the MSM top-level model. \textit{Max(linear-T-scaled)} is the best performing individual measure, indicating the importance of temperature scaling for calibration and OOD detection \cite{liang2020enhancing}. \textit{Global FRE} is the worst performing individual feature. Some of the measures have high AUROC values but relatively low TPR@1\%FPR (\textit{Feature entropy} and \textit{Max(kNN)}).

\subsection{Performance of COOD}
\begin{figure*}
    \centering
    \begin{subfigure}{0.33\linewidth}
        \includegraphics[width=1.0\textwidth]{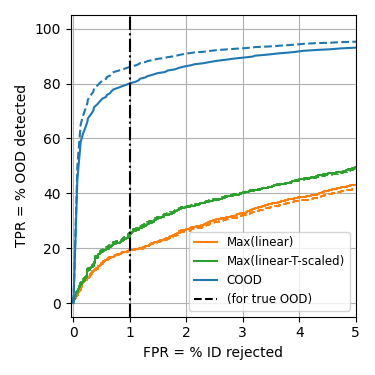}
        \caption{MSM top-level}
        \label{fig:msm_top_roc}
    \end{subfigure}
    \begin{subfigure}{0.33\linewidth}
        \includegraphics[width=1.0\textwidth]{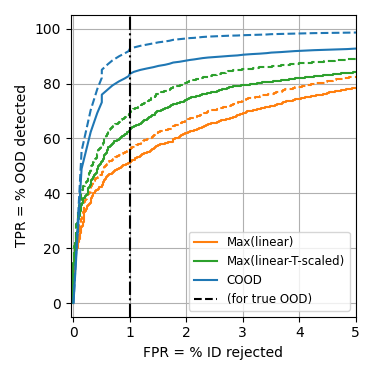}
        \caption{Norwegian vertebrates}
        \label{fig:nor_roc}
    \end{subfigure}
    \begin{subfigure}{0.33\linewidth}
        \includegraphics[width=1.0\textwidth]{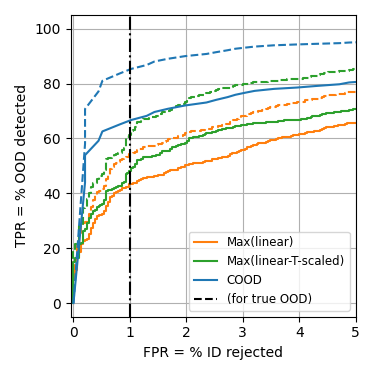}
        \caption{iNaturalist 2018}
        \label{fig:inat_roc}
    \end{subfigure}
    \caption{ROC analysis for OOD detection. The 1\% FPR operating point is indicated by the vertical dot-dashed line. ROC curves are shown for ID-correct vs rest (solid line) and for true OOD (ID-incorrect*  excluded; dashed). At the 1\% FPR operating point \textit{COOD} significantly outperforms both the best single individual measurement \textit{Max(linear-T-scaled)} and the baseline \textit{Max(linear)} for all datasets. Note that the ROC plot is adapted to show the 0-5\% FPR range.}
\hfill
\end{figure*}

For these first analyses, the combined classifier was trained with ID-correct vs rest (ID-incorrect-high, ID-incorrect, OOD-*), preventing the classifier from getting confused by noisy labels from ID-incorrect images when they would have been included as negative (ID) cases. \Cref{fig:msm_top_roc} shows the ROC analysis for the MSM top-level model. COOD outperforms both the baseline (\textit{Max(linear)}) and the best individual measure (\textit{Max(linear-T-scaled)}) by a large margin. When the ID-incorrect* images are excluded from the analysis (\Cref{sec:ref_def}) COOD detects 85.8\% of the OOD images. \Cref{tab:reject_table} shows per ID/OOD category how many images are rejected by COOD and by \textit{Max(linear)}. Note that where \textit{Max(linear)} detects 0\% of the ID-incorrect-high category, COOD detects 22.1\%. Also note that COOD score statistics (mean, stdev, and median) differ between different ID categories. 

\Cref{fig:nor_roc} shows ROC analysis for the Norwegian vertebrates dataset. COOD again outperforms both baseline \textit{Max(linear)} and the best individual measure \textit{Max(linear-T-scaled)} by a large margin. When the ID-incorrect* images are excluded from the analysis COOD detects 94.6\% of the OOD images. \Cref{tab:reject_table} shows per ID/OOD category how many images are rejected by \textit{COOD} and by \textit{Max(linear)}. \textit{COOD} has significantly higher OOD detection percentages than the baseline for both anomaly detection (OOD-far) and novel class detection (OOD-mid and OOD-near). 

\Cref{fig:inat_roc} shows ROC analysis for the iNaturalist 2018 OOD model. COOD again outperforms both baseline \textit{Max(linear)} and the best individual measure \textit{Max(linear-T-scaled)} by a large margin, but performance is lower than MSM top-level and Norwegian vertebrates overall. \Cref{tab:reject_table} shows that for this model the \textit{ID-incorrect*} category is barely detected by any of the measures. This is a combination of the relatively low classifier accuracy and the effect of reference (\Cref{sec:ref_def}, \Cref{tab:inat_cat_ood_app}), see also the discussion.

\begin{table*}
    \centering
    \begin{tabular}{p{\dimexpr.35\linewidth-2\tabcolsep-1.3333\arrayrulewidth}p{\dimexpr.11\linewidth-2\tabcolsep-1.3333\arrayrulewidth}p{\dimexpr.15\linewidth-2\tabcolsep-1.3333\arrayrulewidth}p{\dimexpr.19\linewidth-2\tabcolsep-1.3333\arrayrulewidth}p{\dimexpr.19\linewidth-2\tabcolsep-1.3333\arrayrulewidth}}\hline
    Dataset, category                      &   Number of images &   \% OOD detected - \textit{COOD} &   \% OOD detected - \textit{Max(linear)} & \textit{COOD} - mean, stdev, median              \\
    \hline
   \textbf{MSM top-level}&&&&\\
     ID-correct                     &              10187 &                     1   &                            1   & 0.082, 0.162, 0.010 \\
 ID-incorrect-high: \ensuremath{>}80\% \& TD\ensuremath{>}4 &                276 &                    22.5 &                            0   & 0.514, 0.291, 0.550 \\
 ID-incorrect: \ensuremath{<}80\% $\vert$ TD\ensuremath{<}=4     &                411 &                    30.2 &                           32.8 & 0.664, 0.206, 0.690 \\
 OOD-far: ImageNet-Non-Organism        &               5792 &                    85.8 &                           19.3 & 0.908, 0.191, 0.990 \\
 OOD-far: ImageNet-Cars       &                199 &                    97   &                           12.1 & 0.980, 0.079, 1.000 \\
\hline

    \textbf{Norwegian vertebrates}&&&&\\
     ID-correct                           &              11153 &                     1   &                            1   & 0.107, 0.194, 0.020 \\
 ID-incorrect-high: \ensuremath{>}80\% \& TD\ensuremath{>}4       &                941 &                    15.1 &                            8.1 & 0.487, 0.309, 0.460 \\
 ID-incorrect: \ensuremath{<}80\% $\vert$ TD\ensuremath{<}=4           &                833 &                    36.3 &                           35.3 & 0.718, 0.266, 0.800 \\
 OOD-far: ImageNet-Non-Organism              &               5706 &                    94.6 &                           55.3 & 0.976, 0.073, 1.000 \\
 OOD-mid: Norwegian non-vertebrates   &               5617 &                    91.3 &                           57.7 & 0.965, 0.093, 1.000 \\
 OOD-near: Non-Norwegian vertebrates&                230 &                    65.7 &                           28.7 & 0.820, 0.281, 0.960 \\
\hline
    \textbf{iNaturalist 2018}&&&&\\
     ID-correct                     &               2928 &                     0.9$\dagger$ &                            1   & 0.285, 0.255, 0.210 \\
 ID-incorrect-high: \ensuremath{>}80\% \& TD\ensuremath{>}4 &                193 &                     0   &                            0   & 0.348, 0.216, 0.330 \\
 ID-incorrect: \ensuremath{<}80\% $\vert$ TD\ensuremath{<}=4     &               1787 &                    11.9 &                           12.2 & 0.714, 0.206, 0.750 \\
 OOD-far: ImageNet-Non-Organism &               5738 &                    83.3 &                           54.3 & 0.965, 0.096, 1.000 \\
\hline

    \end{tabular}
    \caption{Results on validation set in terms of \% OOD detected at the 1\%FPR operating point. \textit{COOD} outperforms \textit{Max(linear)} for most of the categories. $\dagger$ not exactly 1.0 due to the ROC curve being discrete}
    \label{tab:reject_table}
\end{table*}

\subsection{Effect of reference}
Classification of taxa on the (sub)species level is a fine-grained task, and it can be difficult to distinguish highly related species. Often there are other reasons why images are incorrectly classified such as poor image quality (out-of-focus, subject too small, presence of other subjects, etc.). In those ID-incorrect cases, it can be expected that their COOD scores differ from ID-correct images. In this section, we evaluate the explicit categorisation of ID-incorrect images both on the definition of the OOD model and the evaluation of the results. We defined 4 different settings (\Cref{tab:exp_settings}) and evaluated its combinations, excluding the logically incompatible (\textit{Classifier definition}=Multi-class, \textit{Multiclass score}=ID-correct) pair, giving 16 combinations in total.

\begin{table*}
    \centering
    \begin{tabularx}{\linewidth}{p{\dimexpr.20\linewidth-2\tabcolsep-1.3333\arrayrulewidth}p{\dimexpr.80\linewidth-2\tabcolsep-1.3333\arrayrulewidth}}
        \toprule
        \textit{Classifier definition} & (1) \textbf{Multi-class}: use 4 categories: ID-correct, ID-incorrect-high, ID-incorrect, OOD*, (2) \textbf{Correct vs rest}: use 2 categories (ID-correct) vs (ID-incorrect-high, ID-incorrect, OOD*), (3) \textbf{ID vs OOD}: 2 categories (ID-correct, ID-incorrect-high, ID-incorrect) vs (OOD*)\\
        \midrule
        \textit{Exclude incorrect from ROC}  & Exclude ID-incorrect-* when evaluating ROC \textbf{Yes / No}\\
        \midrule
        \textit{ROC truth} & Reference when computing ROC (1) \textbf{ID vs OOD}: idem as classifier definition (2) \textbf{not(ID-correct)}: (ID-correct) vs (ID-incorrect-high, ID-incorrect, OOD*) \\
        \midrule
        \textit{Multiclass score} & when Classifier definition=Multi-class how the combined OD measure is computed (1) \textbf{ID-correct}: take 1 - probability of ID-correct (2) \textbf{ID}: take 1 – (sum of probabilities of ID-*)\\
        \bottomrule
    \end{tabularx}
    \caption{Definitions used to determine the influence of several settings on OOD model performance and practical applicability}
    \label{tab:exp_settings}
\end{table*}

\Cref{tab:main_cat_ood} shows the results for selected combinations of the different settings (see \Cref{sec:effect_ref_app} for all results for all datasets). The table is sorted by ascending TPR@1\%FPR. Next to the two main evaluation measures we also include \% ID-incorrect* rejected, indicating how many of the images incorrectly classified by the original task were rejected.

\begin{table*}
    \centering
    \begin{tabular}{
        p{\dimexpr.18\linewidth-2\tabcolsep-1.3333\arrayrulewidth}
        p{\dimexpr.11\linewidth-2\tabcolsep-1.3333\arrayrulewidth}
        p{\dimexpr.15\linewidth-2\tabcolsep-1.3333\arrayrulewidth}
        p{\dimexpr.13\linewidth-2\tabcolsep-1.3333\arrayrulewidth}
        p{\dimexpr.09\linewidth-2\tabcolsep-1.3333\arrayrulewidth}
        p{\dimexpr.10\linewidth-2\tabcolsep-1.3333\arrayrulewidth}
        p{\dimexpr.10\linewidth-2\tabcolsep-1.3333\arrayrulewidth}
        p{\dimexpr.15\linewidth-2\tabcolsep-1.3333\arrayrulewidth}
        }
\toprule
 Classifier definition   & Exclude incorrect from ROC   & ROC truth       & Multiclass score   &   AUROC &   TPR @1\%FPR &   \% ID-incorrect* rejected &   \% ID-incorrect* rejected - min \\
\midrule
 ID-correct vs rest      & no                           & ID vs OOD       & ID                 &    98.2 &        78   &                        9.8 &                              9.8 \\
 ID-correct vs rest      & no                           & not(ID-correct) & ID                 &    98.4 &        80.6 &                       27.4 &                             27.4 \\
 ID vs OOD               & no                           & ID vs OOD       & ID                 &    98.7 &        84.4 &                        5.4 &                              5.4 \\
 Multiclass              & no                           & ID vs OOD       & ID                 &    98.7 &        84.6 &                        5.8 &                              5.8 \\
 ID-correct vs rest      & yes                          & not(ID-correct) & ID                 &    98.8 &        86.5 &                       27.4 &                             27.4 \\
 Multiclass              & yes                          & not(ID-correct) & ID-correct         &    98.8 &        87   &                       26.6 &                              7.3 \\
 Multiclass              & yes                          & not(ID-correct) & ID                 &    98.9 &        87   &                        8   &                              8   \\
 ID vs OOD               & yes                          & not(ID-correct) & ID                 &    98.9 &        87.4 &                        7.6 &                              7.6 \\
\bottomrule
\end{tabular}
    \caption{MSM top-level: selected results for combination of different settings}
    \label{tab:main_cat_ood}
\end{table*}

The best performing combination of settings were all with \textit{Exclude incorrect from ROC}=yes indicating that the \textit{ID-incorrect} samples have overlap with OOD samples, and considering them is important for setting a good operating (decision) threshold. While using \textit{Classifier definition}=\textit{ID vs OOD} results in higher TPR@1\%FPR, the percentage of \textit{ID-incorrect*} rejected is lower. Therefore, we choose as optimal setting (\textit{Classifier definition}=Multi-class, \textit{Exclude incorrect from ROC}=yes, \textit{ROC truth}=not(ID-correct), \textit{Multiclass score}=ID-correct). The multi-class classifier allows not only to distinguish between ID and OD but also between different OOD and ID-incorrect* categories. By reclassifying rejected images based on their Multiclass OOD label the \% ID-incorrect* rejected can be changed from 26.6\% to 7.3\% (\textit{\% ID-incorrect* rejected - min}), depending on e.g. the requirements of an (end-user) application. 

\subsection{SHAP analysis of OOD models}
The Multi-class OOD classifiers were analysed using SHAP analysis \cite{vstrumbelj2014explaining} to determine which individual OOD measures are important contributors – alone or in the context of others – to the COOD model. 

\begin{figure}
    \centering
    \includegraphics[width=1.00\linewidth]
    {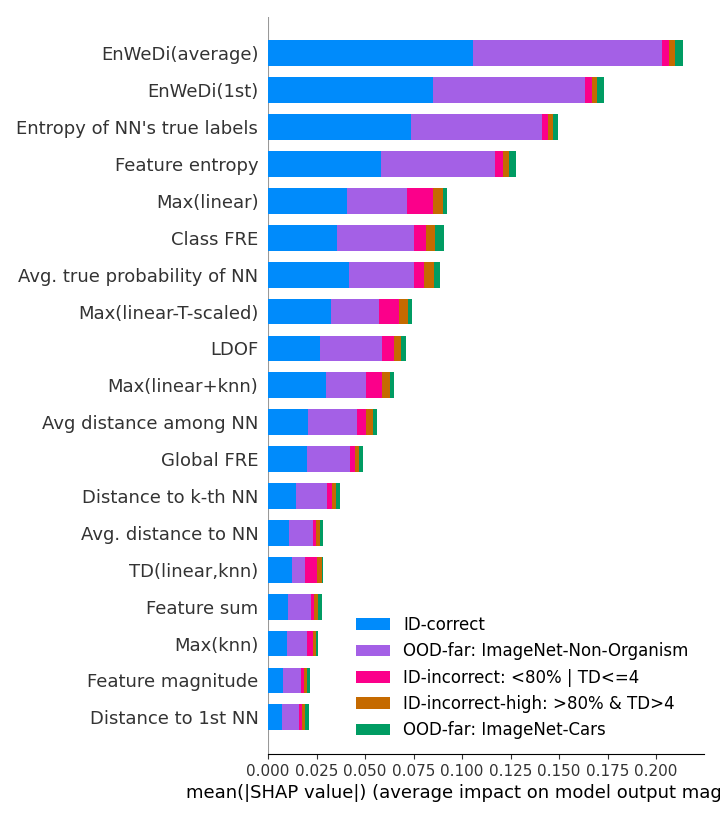}
    \caption{SHAP analysis of MSM top-level showing OOD features most contributing to COOD}
    \label{fig:shap_msm_top}
\end{figure}

\Cref{fig:shap_msm_top} shows the SHAP analysis of the MSM top-level OOD model. For MSM top-level (\Cref{fig:shap_msm_top}) the two \textit{EnWeDi} measures are most contributing, followed by the \textit{Entropy of NN's true class} and \textit{Feature entropy}. \Cref{fig:shap_nor} shows that for the Norwegian vertebrates dataset \textit{Max(linear-T-scaled)} is the most contributing, followed by  \textit{Max(linear+knn)}, \textit{Max(linear)} and the two \textit{EnWeDi} measures. \Cref{fig:inat_shap_class} shows that for iNaturalist 2018 \textit{Max(linear-T-scaled)} is the most contributing, followed by \textit{Max(linear)} and the \textit{Avg. true probability of NN}. The SHAP plots of the contributing measures per OOD class (\Cref{sec:shap_class_app}) show that different measures are important for each class. For example, to detect \textit{ID-incorrect-high} cases \textit{Average true probability of NN} is an important contributor. 

\section{Discussion}

Comparing the results of individual OOD measures and their SHAP contributions shows that OOD measures that individually perform relatively weak can still be important features when used in a combination through a classifier. These effects are well known from machine learning classification literature \cite{mohandes2018classifiers}. It shows that it could be a better strategy to develop a diverse set of relatively weak OOD measures which cancel out each other’s weaknesses than to try to develop a single state-of-the-art OOD measure. The presented framework for combining OOD measures is easily extendable with others, either existing ones or newly developed ones that tackle specific weaknesses of other measures.

A consideration when developing OOD measures is their computation time. Many of the most contributing measures presented in this paper were based on a kNN model. Although kNN models are relatively expensive, they are very powerful, public efficient implementations are available, and they allow visual inspection of the neighbours which can give insight into (apparent) errors \cite{papernot2018deep}. A benefit of using a relatively large number of, possibly weak, individual OOD measures is that it might be possible to omit measures that have prohibitively large computation requirements (time and/or memory) during training and inference. It should be well tested then that omitting a specific OOD feature does not decrease the performance for a specific OOD class.

In this work we present two novel OOD measures based on the discrepancy between linear and kNN predictions. According to SHAP analyses both are important contributing features for the novel class detection task. We hypothesise that these features work because in the OOD part of the feature space decision boundaries are not well defined and the linear and kNN decision boundary disagree more often. It is known that applying a non-linear classifier, such as kNN, to the feature space of a deep neural network can improve over the standard linear classification model \cite{notley2018examining} indicating that the linear model contains different and sometimes sub-optimal information.  

As far as we are aware previous literature does not explicitly deal with images that were incorrectly identified by the original task's – non-OOD – classifier. We found that using an explicit categorisation is important both for training the combined OOD classifier and for interpreting the results. For detecting \textit{ID-incorrect*} cases, it is helpful if the original classifier's accuracy is high, but for less accurate original models the method still works when the correct reference settings are chosen. Many of the \textit{ID-incorrect} images are incorrectly classified because they are poor quality or deviate from other samples and could be considered as ID anomalies and could share similarities with OOD anomalies (out-of-domain; \Cref{fig:false_positive_example}). In interactive applications, it could be very useful to not only flag true OOD images to the user but also ID images which the OOD model thinks are incorrectly classified by the original model (\Cref{fig:id-incorrect-high_example}). 

\begin{figure}
    \centering
    \begin{subfigure}[t]{0.48\linewidth}
        \includegraphics[width=1.0\textwidth]{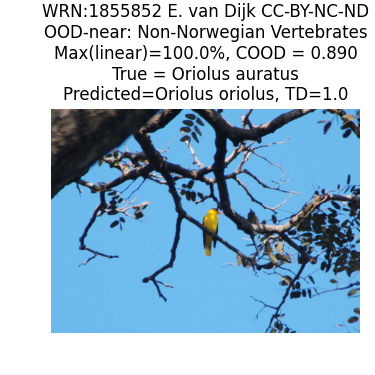}
        \caption{Norwegian vertebrates: successful detection of closely related novel class}
        \label{fig:novel_class}
    \end{subfigure}
    \begin{subfigure}[t]{0.48\linewidth}
        \includegraphics[width=1.0\textwidth]{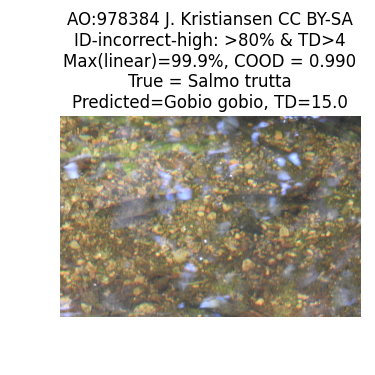}
        \caption{Norwegian vertebrates: successful detection of highly confident wrong prediction}
        \label{fig:id-incorrect-high_example}
    \end{subfigure}
    \begin{subfigure}[t]{0.48\linewidth}
        \includegraphics[width=1.0\textwidth]{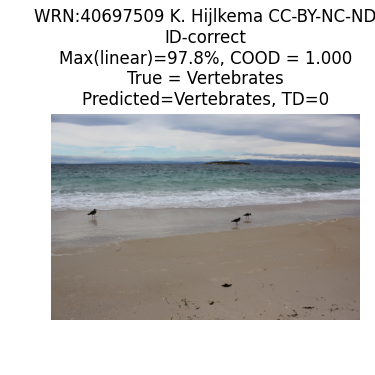}
        \caption{MSM top-level: 'false positive'. Could also be seen as low quality image for species recognition}
        \label{fig:false_positive_example}
    \end{subfigure}
    \begin{subfigure}[t]{0.48\linewidth}
        \includegraphics[width=1.0\textwidth]{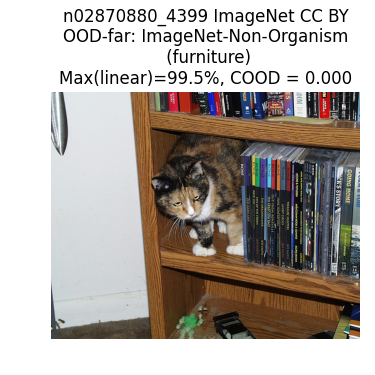}
        \caption{MSM top-level: 'false negative'. Result of ImageNet's label being 'furniture' and not 'organism'}
        \label{fig:false_negative_example}
    \end{subfigure}
    \caption{Example images}
    \label{fig:example_figures}
\end{figure}
 
\textit{ID-correct} images can look similar to OOD (ImageNet) images as well (\Cref{fig:false_positive_example}). Vice-versa OOD images can look like ID images (\Cref{fig:false_negative_example}), in this case due to a label inconsistency in ImageNet. While we could have further refined the definition of OOD for ImageNet, label inconsistencies and multi-subject images make this a problematic endeavour. Choosing datasets that have no domain overlap with biodiversity (\textit{OOD-far: ImageNet-Cars}) is an option, but leads to very high OOD performances. We prefer the insight that can be gained from the examples of FP and FN cases in harder problems (\Cref{app:example_images}). We conjecture that class overlap is a reason that the Norwegian vertebrates model detects \textit{OOD-far: ImageNet-Non-Organism} images better than MSM top-level. The MSM top-level has broad classes and relatively many images look similar to ImageNet, while the Norwegian vertebrates model is specific for vertebrates and has less variation. 


Extending the list of individual metrics with other existing and novel individual measures could further improve performance, the same for trying different classifiers for the \textit{COOD} model (MLP, SVM, etc.). Ablation and optimisation studies to investigate the influence of parameters such as $k$ (used in kNN search), the effect of the kNN distance metric, etc. are helpful and will be included in an extended version of this work. We evaluated on biodiversity datasets, but expect that a modified version will work well on other datasets (e.g. non-hierarchical) too. Better dealing with class/domain overlap in OOD evaluation is an important topic raised by others too \cite{hendrycks2019, bitterwolf2023out}. Finally, we expect that feature spaces with improved properties, as a result of alternative training methods such as supervised contrastive loss \cite{sun2022out}, deeper features \cite{papernot2018deep}, or different neural architectures such as Vision Transformers \cite{fort2021exploring} can improve the performance further.

\section{Conclusion}

Our paper presents three contributions to the topic of anomaly and novel class (\Cref{fig:novel_class}) detection: (1) a learned combination of existing and novel individual OOD measures outperforms significantly the best individual OOD measures (2) the discrepancy between linear and kNN predictions forms an important novel individual OOD measure for novel class detection, specifically we exploit the hierarchical class structure, (3) explicitly defining how to deal with in-distribution but incorrect predictions is important for consistent analysis and application in practice.

\section{Acknowledgements}

We would like to thank the citizen scientists contributing to and the experts validating images in the databases of Observation.org, Artsobservasjoner.no, Artportalen.se, Arter.dk, iRecord.org.uk, and iNaturalist.org for their efforts without which this work would not have been possible. 

This research was supported by the EU Horizon Europe projects MAMBO programme under grant agreement No.101060639.

\newpage
{
    \small
    \bibliographystyle{ieeenat_fullname}
    \bibliography{main}

\begin{thebibliography}{38}
\providecommand{\natexlab}[1]{#1}
\providecommand{\url}[1]{\texttt{#1}}
\expandafter\ifx\csname urlstyle\endcsname\relax
  \providecommand{\doi}[1]{doi: #1}\else
  \providecommand{\doi}{doi: \begingroup \urlstyle{rm}\Url}\fi

\bibitem[Abdar et~al.(2021)Abdar, Pourpanah, Hussain, Rezazadegan, Liu, Ghavamzadeh, Fieguth, Cao, Khosravi, Acharya, et~al.]{abdar2021review}
Moloud Abdar, Farhad Pourpanah, Sadiq Hussain, Dana Rezazadegan, Li Liu, Mohammad Ghavamzadeh, Paul Fieguth, Xiaochun Cao, Abbas Khosravi, U~Rajendra Acharya, et~al.
\newblock A review of uncertainty quantification in deep learning: Techniques, applications and challenges.
\newblock \emph{Information fusion}, 76:\penalty0 243--297, 2021.

\bibitem[Amodei et~al.(2016)Amodei, Olah, Steinhardt, Christiano, Schulman, and Man{\'{e}}]{AmodeiOSCSM16}
Dario Amodei, Chris Olah, Jacob Steinhardt, Paul~F. Christiano, John Schulman, and Dan Man{\'{e}}.
\newblock Concrete problems in {AI} safety.
\newblock \emph{CoRR}, abs/1606.06565, 2016.

\bibitem[Anubha~Pearline et~al.(2019)Anubha~Pearline, Sathiesh~Kumar, and Harini]{anubha2019study}
S Anubha~Pearline, V Sathiesh~Kumar, and S Harini.
\newblock A study on plant recognition using conventional image processing and deep learning approaches.
\newblock \emph{Journal of Intelligent \& Fuzzy Systems}, 36\penalty0 (3):\penalty0 1997--2004, 2019.

\bibitem[Aodha(2018)]{inat2018_pretrained_model_github_repo}
Oisin~Mac Aodha.
\newblock {iNaturalist Competition 2018 Training Code}.
\newblock \url{https://github.com/macaodha/inat_comp_2018}, 2018.

\bibitem[Bergman et~al.(2020)Bergman, Cohen, and Hoshen]{bergman2020deep}
Liron Bergman, Niv Cohen, and Yedid Hoshen.
\newblock Deep nearest neighbor anomaly detection.
\newblock \emph{arXiv preprint arXiv:2002.10445}, 2020.

\bibitem[Bitterwolf et~al.(2023)Bitterwolf, Müller, and Hein]{bitterwolf2023out}
Julian Bitterwolf, Maximilian Müller, and Matthias Hein.
\newblock In or out? fixing imagenet out-of-distribution detection evaluation, 2023.

\bibitem[Deng et~al.(2009)Deng, Dong, Socher, Li, Li, and Fei-Fei]{ImageNet_dataset}
Jia Deng, Wei Dong, Richard Socher, Li-Jia Li, Kai Li, and Li Fei-Fei.
\newblock Imagenet: A large-scale hierarchical image database.
\newblock In \emph{2009 IEEE Conference on Computer Vision and Pattern Recognition}, pages 248--255, 2009.

\bibitem[Diwan et~al.(2021)Diwan, Choubey, Hota, Goyal, Jamal, Shukla, and Tiwari]{diwan2021feature}
Tarun~Dhar Diwan, Siddartha Choubey, HS Hota, SB Goyal, Sajjad~Shaukat Jamal, Piyush~Kumar Shukla, and Basant Tiwari.
\newblock Feature entropy estimation (fee) for malicious iot traffic and detection using machine learning.
\newblock \emph{Mobile Information Systems}, 2021:\penalty0 1--13, 2021.

\bibitem[Douze et~al.(2024)Douze, Guzhva, Deng, Johnson, Szilvasy, Mazar{\'e}, Lomeli, Hosseini, and J{\'e}gou]{douze2024faiss}
Matthijs Douze, Alexandr Guzhva, Chengqi Deng, Jeff Johnson, Gergely Szilvasy, Pierre-Emmanuel Mazar{\'e}, Maria Lomeli, Lucas Hosseini, and Herv{\'e} J{\'e}gou.
\newblock The faiss library.
\newblock \emph{arXiv preprint arXiv:2401.08281}, 2024.

\bibitem[Feng et~al.(2022)Feng, Harakeh, Waslander, and Dietmayer]{di9525313}
Di Feng, Ali Harakeh, Steven~L. Waslander, and Klaus Dietmayer.
\newblock A review and comparative study on probabilistic object detection in autonomous driving.
\newblock \emph{IEEE Transactions on Intelligent Transportation Systems}, 23\penalty0 (8):\penalty0 9961--9980, 2022.

\bibitem[Fort et~al.(2021)Fort, Ren, and Lakshminarayanan]{fort2021exploring}
Stanislav Fort, Jie Ren, and Balaji Lakshminarayanan.
\newblock Exploring the limits of out-of-distribution detection.
\newblock \emph{Advances in Neural Information Processing Systems}, 34:\penalty0 7068--7081, 2021.

\bibitem[Ganaie et~al.(2022)Ganaie, Hu, Malik, Tanveer, and Suganthan]{ganaie2022ensemble}
Mudasir~A Ganaie, Minghui Hu, AK Malik, M Tanveer, and PN Suganthan.
\newblock Ensemble deep learning: A review.
\newblock \emph{Engineering Applications of Artificial Intelligence}, 115:\penalty0 105151, 2022.

\bibitem[{Gangireddy}(2023)]{gangireddy2023}
Rajesh {Gangireddy}.
\newblock Knowing the unknown : Open-world recognition for biodiversity datasets.
\newblock Master's thesis, University of Twente, 2023.

\bibitem[Hendrycks and Gimpel(2018)]{hendrycks2018baseline}
Dan Hendrycks and Kevin Gimpel.
\newblock A baseline for detecting misclassified and out-of-distribution examples in neural networks, 2018.

\bibitem[Hendrycks et~al.(2018)Hendrycks, Mazeika, and Dietterich]{hendrycks2018deep}
Dan Hendrycks, Mantas Mazeika, and Thomas Dietterich.
\newblock Deep anomaly detection with outlier exposure.
\newblock \emph{arXiv preprint arXiv:1812.04606}, 2018.

\bibitem[Hendrycks et~al.(2019)Hendrycks, Basart, Mazeika, Mostajabi, Steinhardt, and Song]{hendrycks2019}
Dan Hendrycks, Steven Basart, Mantas Mazeika, Mohammadreza Mostajabi, Jacob Steinhardt, and Dawn Song.
\newblock A benchmark for anomaly segmentation.
\newblock \emph{CoRR}, abs/1911.11132, 2019.

\bibitem[Ho(1995)]{ho1995random}
Tin~Kam Ho.
\newblock Random decision forests.
\newblock In \emph{Proceedings of 3rd international conference on document analysis and recognition}, pages 278--282. IEEE, 1995.

\bibitem[Horn et~al.(2018)Horn, Aodha, Song, Cui, Sun, Shepard, Adam, Perona, and Belongie]{inat_dataset}
Grant Horn, Oisin Aodha, Yang Song, Yin Cui, Chen Sun, Alex Shepard, Hartwig Adam, Pietro Perona, and Serge Belongie.
\newblock The inaturalist species classification and detection dataset.
\newblock In \emph{2018 IEEE/CVF Conference on Computer Vision and Pattern Recognition (CVPR)}, pages 8769--8778, 2018.

\bibitem[Liang et~al.(2020)Liang, Li, and Srikant]{liang2020enhancing}
Shiyu Liang, Yixuan Li, and R. Srikant.
\newblock Enhancing the reliability of out-of-distribution image detection in neural networks, 2020.

\bibitem[Liu et~al.(2021)Liu, Shen, He, Zhang, Xu, Yu, and Cui]{liu2021towards}
Jiashuo Liu, Zheyan Shen, Yue He, Xingxuan Zhang, Renzhe Xu, Han Yu, and Peng Cui.
\newblock Towards out-of-distribution generalization: A survey.
\newblock \emph{arXiv preprint arXiv:2108.13624}, 2021.

\bibitem[Liu et~al.(2020)Liu, Wang, Owens, and Li]{Liu2020EnergybasedOD}
Weitang Liu, Xiaoyun Wang, John~Douglas Owens, and Yixuan Li.
\newblock Energy-based out-of-distribution detection.
\newblock \emph{ArXiv}, abs/2010.03759, 2020.

\bibitem[Mohandes et~al.(2018)Mohandes, Deriche, and Aliyu]{mohandes2018classifiers}
Mohamed Mohandes, Mohamed Deriche, and Salihu~O Aliyu.
\newblock Classifiers combination techniques: A comprehensive review.
\newblock \emph{IEEE Access}, 6:\penalty0 19626--19639, 2018.

\bibitem[Ndiour et~al.(2020)Ndiour, Ahuja, and Tickoo]{ndiour2020out}
Ibrahima Ndiour, Nilesh Ahuja, and Omesh Tickoo.
\newblock Out-of-distribution detection with subspace techniques and probabilistic modeling of features.
\newblock \emph{arXiv preprint arXiv:2012.04250}, 2020.

\bibitem[Nguyen et~al.(2015)Nguyen, Yosinski, and Clune]{nguyen2015deep}
Anh Nguyen, Jason Yosinski, and Jeff Clune.
\newblock Deep neural networks are easily fooled: High confidence predictions for unrecognizable images.
\newblock In \emph{Proceedings of the IEEE conference on computer vision and pattern recognition}, pages 427--436, 2015.

\bibitem[Notley and Magdon-Ismail(2018)]{notley2018examining}
Stephen Notley and Malik Magdon-Ismail.
\newblock Examining the use of neural networks for feature extraction: A comparative analysis using deep learning, support vector machines, and k-nearest neighbor classifiers.
\newblock \emph{arXiv preprint arXiv:1805.02294}, 2018.

\bibitem[Papernot and McDaniel(2018)]{papernot2018deep}
Nicolas Papernot and Patrick McDaniel.
\newblock Deep k-nearest neighbors: Towards confident, interpretable and robust deep learning.
\newblock \emph{arXiv preprint arXiv:1803.04765}, 2018.

\bibitem[Roady et~al.(2019)Roady, Hayes, Kemker, Gonzales, and Kanan]{abs-1910-14034}
Ryne Roady, Tyler~L. Hayes, Ronald Kemker, Ayesha Gonzales, and Christopher Kanan.
\newblock Are out-of-distribution detection methods effective on large-scale datasets?
\newblock \emph{CoRR}, abs/1910.14034, 2019.

\bibitem[Ruff et~al.(2021)Ruff, Kauffmann, Vandermeulen, Montavon, Samek, Kloft, Dietterich, and M{\"u}ller]{ruff2021unifying}
Lukas Ruff, Jacob~R Kauffmann, Robert~A Vandermeulen, Gr{\'e}goire Montavon, Wojciech Samek, Marius Kloft, Thomas~G Dietterich, and Klaus-Robert M{\"u}ller.
\newblock A unifying review of deep and shallow anomaly detection.
\newblock \emph{Proceedings of the IEEE}, 109\penalty0 (5):\penalty0 756--795, 2021.

\bibitem[Schermer and Hogeweg(2018)]{schermer2018supporting}
Maarten Schermer and Laurens Hogeweg.
\newblock Supporting citizen scientists with automatic species identification using deep learning image recognition models.
\newblock \emph{Biodiversity Information Science and Standards}, 2018.

\bibitem[{\v{S}}trumbelj and Kononenko(2014)]{vstrumbelj2014explaining}
Erik {\v{S}}trumbelj and Igor Kononenko.
\newblock Explaining prediction models and individual predictions with feature contributions.
\newblock \emph{Knowledge and information systems}, 41:\penalty0 647--665, 2014.

\bibitem[Sun et~al.(2022)Sun, Ming, Zhu, and Li]{sun2022out}
Yiyou Sun, Yifei Ming, Xiaojin Zhu, and Yixuan Li.
\newblock Out-of-distribution detection with deep nearest neighbors.
\newblock In \emph{International Conference on Machine Learning}, pages 20827--20840. PMLR, 2022.

\bibitem[Szegedy et~al.(2016)Szegedy, Vanhoucke, Ioffe, Shlens, and Wojna]{inceptionv3_paper}
Christian Szegedy, Vincent Vanhoucke, Sergey Ioffe, Jon Shlens, and Zbigniew Wojna.
\newblock Rethinking the inception architecture for computer vision.
\newblock In \emph{2016 IEEE Conference on Computer Vision and Pattern Recognition (CVPR)}, pages 2818--2826, 2016.

\bibitem[Tajwar et~al.(2021)Tajwar, Kumar, Xie, and Liang]{abs-2109-05554}
Fahim Tajwar, Ananya Kumar, Sang~Michael Xie, and Percy Liang.
\newblock No true state-of-the-art? {OOD} detection methods are inconsistent across datasets.
\newblock \emph{CoRR}, abs/2109.05554, 2021.

\bibitem[Tan and Le(2019)]{tan2019efficientnet}
Mingxing Tan and Quoc Le.
\newblock Efficientnet: Rethinking model scaling for convolutional neural networks.
\newblock In \emph{International conference on machine learning}, pages 6105--6114. PMLR, 2019.

\bibitem[Tian et~al.(2021)Tian, Pang, Chen, Singh, Verjans, and Carneiro]{tian2021weakly}
Yu Tian, Guansong Pang, Yuanhong Chen, Rajvinder Singh, Johan~W Verjans, and Gustavo Carneiro.
\newblock Weakly-supervised video anomaly detection with robust temporal feature magnitude learning.
\newblock In \emph{Proceedings of the IEEE/CVF international conference on computer vision}, pages 4975--4986, 2021.

\bibitem[Yousefnezhad et~al.(2021)Yousefnezhad, Hamidzadeh, and Aliannejadi]{yousefnezhad2021ensemble}
Maryam Yousefnezhad, Javad Hamidzadeh, and Mohammad Aliannejadi.
\newblock Ensemble classification for intrusion detection via feature extraction based on deep learning.
\newblock \emph{Soft Computing}, 25\penalty0 (20):\penalty0 12667--12683, 2021.

\bibitem[Zhang et~al.(2009)Zhang, Hutter, and Jin]{zhang2009new}
Ke Zhang, Marcus Hutter, and Huidong Jin.
\newblock A new local distance-based outlier detection approach for scattered real-world data.
\newblock In \emph{Advances in Knowledge Discovery and Data Mining: 13th Pacific-Asia Conference, PAKDD 2009 Bangkok, Thailand, April 27-30, 2009 Proceedings 13}, pages 813--822. Springer, 2009.

\bibitem[Zimmerer et~al.(2022)Zimmerer, Full, Isensee, Jäger, Adler, Petersen, Köhler, Ross, Reinke, Kascenas, Jensen, O’Neil, Tan, Hou, Batten, Qiu, Kainz, Shvetsova, Fedulova, Dylov, Yu, Zhai, Hu, Si, Zhou, Wang, Li, Chen, Zhao, Marimont, Tarroni, Saase, Maier-Hein, and Maier-Hein]{zimmerer9762702}
David Zimmerer, Peter~M. Full, Fabian Isensee, Paul Jäger, Tim Adler, Jens Petersen, Gregor Köhler, Tobias Ross, Annika Reinke, Antanas Kascenas, Bjørn~Sand Jensen, Alison~Q. O’Neil, Jeremy Tan, Benjamin Hou, James Batten, Huaqi Qiu, Bernhard Kainz, Nina Shvetsova, Irina Fedulova, Dmitry~V. Dylov, Baolun Yu, Jianyang Zhai, Jingtao Hu, Runxuan Si, Sihang Zhou, Siqi Wang, Xinyang Li, Xuerun Chen, Yang Zhao, Sergio~Naval Marimont, Giacomo Tarroni, Victor Saase, Lena Maier-Hein, and Klaus Maier-Hein.
\newblock Mood 2020: A public benchmark for out-of-distribution detection and localization on medical images.
\newblock \emph{IEEE Transactions on Medical Imaging}, 41\penalty0 (10):\penalty0 2728--2738, 2022.

\end{thebibliography}
}

\clearpage
\setcounter{page}{1}
\maketitlesupplementary
\section{Appendix}

\subsection{Taxon distance}
\label{sec:td_app}

\vspace{4cm}

\begin{minipage}{\textwidth}
    \captionsetup{type=figure}
    \centering
    \includegraphics[width=1.0\textwidth]{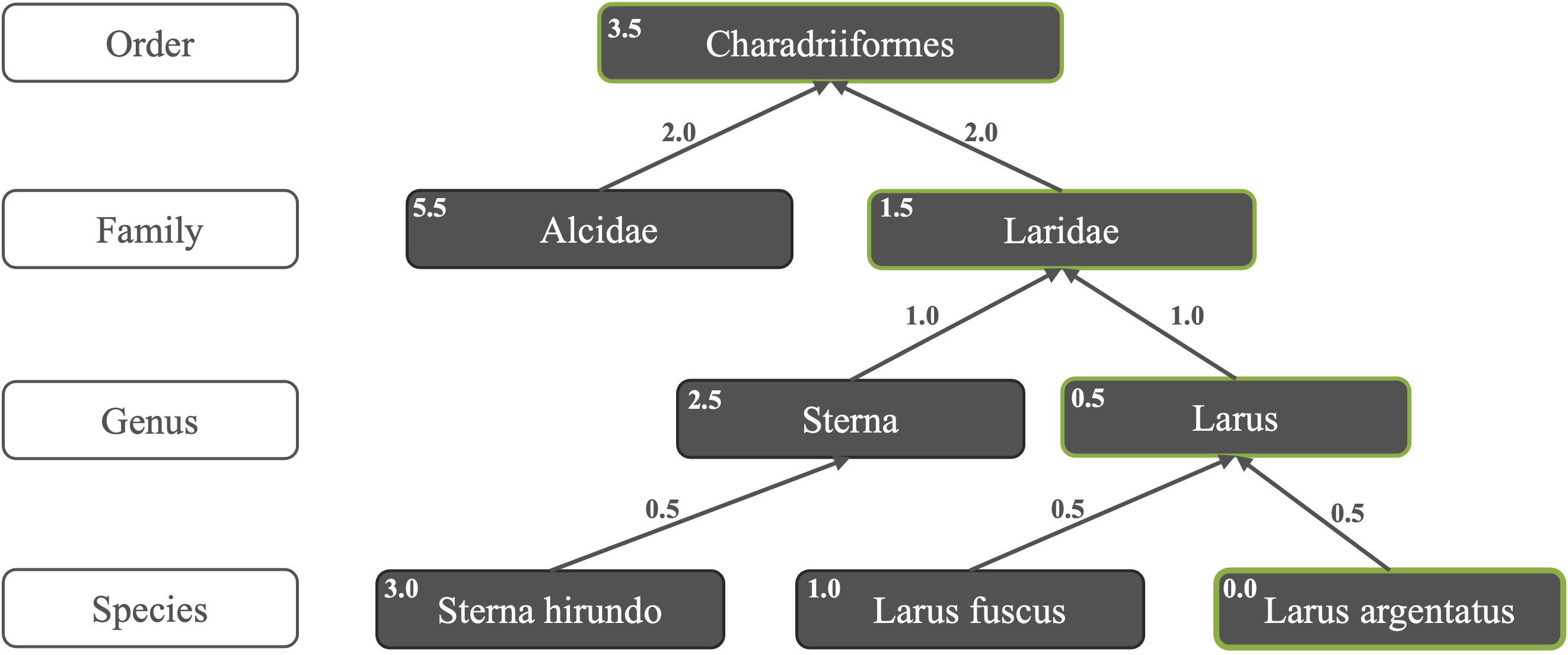}
    \caption{Taxon distance. Shows a part of a hierarchical class tree where taxa (biological classes) make up the nodes. The taxon distance (TD) is defined as the sum of the number of weighted edges between two nodes in the tree (the first node in the example is Larus argentatus). The weights are used to ensure that distances higher in the tree are larger, matching with the concept of broader groups. Per definition the distance between to species in the same genus = 1.0, e.g. TD(Larus argentatus, Larus fuscus) = 1}
    \label{fig:taxon_distance}
\end{minipage}

\clearpage

\subsection{Mathematical definitions}


\vspace{6cm}

\begin{minipage}{\textwidth}
    \captionsetup{type=table}
    \centering
    \begin{tabularx}{\linewidth}{>{\hsize=0.50\hsize}X>{\hsize=1.2\hsize}X>{\hsize=0.3\hsize}X}
        \toprule
        Name & Definition & Symbol \\
        \midrule
        \midrule
        Data point & $x_i$ from dataset $\mathcal{D}$ & $x$\\
        \midrule
        Feature & $f(x)$, (image) feature for $x$ & $f$ \\
        \midrule
        Logits & $g = W*f + b$, where $W$ is the classification weight matrix, $b$ the classification bias, and $g_{c}$ is the linear output for the $c^{th}$ class & $g_{c}$\\
        \midrule
        Temperature scaled linear class probability& $p_{T}(x, T) = \dfrac{e^{g_c(x)/T}}{\sum_{c=1}^{|{\mathcal{C}}|}e^{g_c(x)/T}}$, also known as \textit{SoftMax}, where $\mathcal{C}$ indicates set of classes and $T$ is the temperature scale factor  & $p_{T}$ \\
        \midrule
        Linear class probability& $p_c(x) = p_T(x, 1.0)$  & $p$ \\
        \midrule
        
        Feature distance&$d(f_i, f_j)$, without loss of generality we assume low $d$ means more similar features&$d$\\
        \midrule
        n-th nearest neighbor&$\mathcal{N}_{n}(x)$, as found by kNN search, \newline per definition $\mathcal{N}_1 = {argmin}_{x_j \in X}d(x, x_j)$, assuming $x \notin X$ &$\mathcal{N}_{n}$ \\
        \midrule
        True class&$\mathcal{T}(x)$, gives true class for $x$&$\mathcal{T}(x)$\\
        \midrule
        kNN class histogram&$h_{c, \mathcal{N} }(x) = \Sigma_{n=1}^{k}\textbf{1}[\mathcal{T}(\mathcal{N}_n) = c]$, for all $c \in \mathcal{C}$ where $\textbf{1}[\cdot]$ is the indicator function&$h_{\mathcal{N}}$\\
        \midrule
        kNN class probability vector&$\frac{1}{k}h_{\mathcal{N}}$&$p_{\mathcal{N}}$\\
        \midrule
        Entropy&$H(p) = - \sum_{i=1}^n p_i\log_2(p_i)$, where $p$ is an arbitrary probability  vector&$H$\\
    \end{tabularx}
    \caption{Basic mathematical definitions}
    \label{tab:basic_math_def}
\end{minipage}

\begin{table*}[p]
  \centering
  \begin{tabularx}{\linewidth}{>{\hsize=0.37\hsize}X>{\hsize=1.2\hsize}X>{\hsize=0.3\hsize}X}
  \toprule
    OOD Measure & Definition & Symbol \\
  \midrule\midrule
    \textit{Avg. distance among NN}  & $\frac{1}{k (k -1)}\Sigma_{n}^{k} \Sigma_{n \neq m}^{k}d(f(\mathcal{N}_{m}), {f(\mathcal{N}}_{n}))$  & $\bar{D}$ \\
    \midrule
    \textit{Avg. distance to NN}  & $\frac{1}{k}\Sigma_{n}^{k} d(f, f({\mathcal{N}}_{n}))$ & $\bar{d}$ \\
    \midrule
    \textit{Distance to 1st NN}  & $d(f, f({\mathcal{N}}_{1}))$& $d_1$  \\
    \midrule
    \textit{Distance to k-th NN}  & $d(f, f({\mathcal{N}}_{k}))$  &  $d_k$ \\
    \midrule
    \textit{LDOF} & $\dfrac{\bar{d}}{\bar{D}}$ &   \\
    \midrule
    \textit{Global FRE}  & ${\parallel f -  \hat{\tau}(\tau(f))\parallel}_2  $, where $\tau$, is the forward PCA transformation, and $\hat{\tau}$ is its Moore-Penrose pseudo-inverse
&  \\
    \midrule
    \textit{Class FRE}  & ${\parallel f -  \hat{\tau_{c}}(\tau_{c}(f))\parallel}_2 $ where $\tau_{c}$ is the PCA model for the class predicted by $p$  & \\
    \midrule
    \textit{Max(linear)}  & $max(p)$  & \\
    \midrule
    \textit{Max(knn)}  & $max(p_{\mathcal{N}})$  & \\
    \midrule
    \textit{Max(linear-T-scaled)}  & $max(p_{2.0})$  &   \\
    \midrule
    \textit{Max(linear+kNN)}  & $max((p + p_{\mathcal{N}}) / 2)$ &  \\
    \midrule
    \textit{TD(linear, kNN)}  & See \Cref{fig:taxon_distance}  & TD  \\
    \midrule
    \textit{Entropy of NN’s true class}  & $H(p_{\mathcal{N}})$ & $H_{\mathcal{N}}$\\
    \midrule
    \textit{EnWeDi(1st)}  & $d_{1} \cdot (1+H_{\mathcal{N}}) $  &   \\
    \midrule
    \textit{EnWeDi(average)}  & $\bar{d} \cdot (1+H_{\mathcal{N}}) $ & \\
    \midrule
    \textit{Feature entropy} & $H(f/|f|)$, where $|f|$ defines the number of elements in $f$ &   \\
    \midrule
    \textit{Feature sum}  & $\Sigma_{a}(f_a)$, where $a$ indexes the feature elements  & \\
    \midrule
     \textit{Feature magnitude}  & $\parallel f \parallel_2$ &  \\
    \midrule
    
    \textit{Avg. true probability of NN}  & $\dfrac{1}{k} \Sigma_n^k p_{c=\mathcal{T}(\mathcal{N}_n)}$, where $p_{c=\mathcal{T}(\mathcal{N}_n)}$ is the true class probability for $\mathcal{N}_n$  &   \\
  \end{tabularx}
  \caption{Mathematical definitions of OOD measures. See \Cref{tab:basic_math_def} for basic definitions}
  \label{tab:ood_measures_app}
\end{table*}

\clearpage

\subsection{Effect of reference}
\label{sec:effect_ref_app}
\vspace{2cm}
\begin{minipage}{\textwidth}
    \captionsetup{type=table}
    \centering
    \begin{tabular}{
p{\dimexpr.12\linewidth-2\tabcolsep-1.3333\arrayrulewidth}
p{\dimexpr.09\linewidth-2\tabcolsep-1.3333\arrayrulewidth}
p{\dimexpr.11\linewidth-2\tabcolsep-1.3333\arrayrulewidth}
p{\dimexpr.09\linewidth-2\tabcolsep-1.3333\arrayrulewidth}
p{\dimexpr.09\linewidth-2\tabcolsep-1.3333\arrayrulewidth}
p{\dimexpr.10\linewidth-2\tabcolsep-1.3333\arrayrulewidth}
p{\dimexpr.10\linewidth-2\tabcolsep-1.3333\arrayrulewidth}
p{\dimexpr.10\linewidth-2\tabcolsep-1.3333\arrayrulewidth}
p{\dimexpr.10\linewidth-2\tabcolsep-1.3333\arrayrulewidth}
p{\dimexpr.10\linewidth-2\tabcolsep-1.3333\arrayrulewidth}
}
\toprule
 Classifier definition   & Exclude incorrect from ROC   & ROC truth       & Multi-class score  &   AUROC &   TPR @1\%FPR &   \% ID-incorrect* rejected &   \% ID-incorrect* rejected - min & OOD vs ID-incorrect* accuracy   & OOD vs ID-incorrect* F1   \\
\midrule
 ID-correct vs rest      & no                           & ID vs OOD       & ID                 &    98.2 &        78   &                        9.8 &                              9.8 & -                               & -                         \\\midrule
 Multi-class              & no                           & not(ID-correct) & ID                 &    97   &        78.9 &                        8   &                              8   & -                               & -                         \\\midrule
 ID vs OOD               & no                           & not(ID-correct) & ID                 &    96.9 &        79.6 &                        7.7 &                              7.7 & -                               & -                         \\\midrule
 Multi-class              & no                           & ID vs OOD       & ID-correct         &    98.2 &        79.7 &                        9.9 &                              3.2 & 99.3                            & 83.8                      \\\midrule
 ID-correct vs rest      & no                           & not(ID-correct) & ID                 &    98.4 &        80.6 &                       27.4 &                             27.4 & -                               & -                         \\\midrule
 Multi-class              & no                           & not(ID-correct) & ID-correct         &    98.4 &        81   &                       26.6 &                              7.3 & 98.4                            & 85.8                      \\\midrule
 ID vs OOD               & no                           & ID vs OOD       & ID                 &    98.7 &        84.4 &                        5.4 &                              5.4 & -                               & -                         \\\midrule
 Multi-class              & no                           & ID vs OOD       & ID                 &    98.7 &        84.6 &                        5.8 &                              5.8 & -                               & -                         \\\midrule
 ID-correct vs rest      & yes                          & not(ID-correct) & ID                 &    98.8 &        86.5 &                       27.4 &                             27.4 & -                               & -                         \\\midrule
 ID-correct vs rest      & yes                          & ID vs OOD       & ID                 &    98.8 &        86.5 &                       27.4 &                             27.4 & -                               & -                         \\\midrule
 Multi-class              & yes                          & not(ID-correct) & ID-correct         &    98.8 &        87   &                       26.6 &                              7.3 & 98.4                            & 85.8                      \\\midrule
 Multi-class              & yes                          & ID vs OOD       & ID-correct         &    98.8 &        87   &                       26.6 &                              7.3 & 98.4                            & 85.8                      \\\midrule
 Multi-class              & yes                          & not(ID-correct) & ID                 &    98.9 &        87   &                        8   &                              8   & -                               & -                         \\\midrule
 Multi-class              & yes                          & ID vs OOD       & ID                 &    98.9 &        87   &                        8   &                              8   & -                               & -                         \\\midrule
 ID vs OOD               & yes                          & not(ID-correct) & ID                 &    98.9 &        87.4 &                        7.6 &                              7.6 & -                               & -                         \\\midrule
 ID vs OOD               & yes                          & ID vs OOD       & ID                 &    98.9 &        87.4 &                        7.6 &                              7.6 & -                               & -                         \\
\bottomrule
\end{tabular}
    \caption{Effect of reference: MSM top-level}
    \label{tab:top_cat_ood_app}
\end{minipage}
\begin{table*}
    \centering
    \begin{tabular}{
p{\dimexpr.12\linewidth-2\tabcolsep-1.3333\arrayrulewidth}
p{\dimexpr.09\linewidth-2\tabcolsep-1.3333\arrayrulewidth}
p{\dimexpr.11\linewidth-2\tabcolsep-1.3333\arrayrulewidth}
p{\dimexpr.09\linewidth-2\tabcolsep-1.3333\arrayrulewidth}
p{\dimexpr.09\linewidth-2\tabcolsep-1.3333\arrayrulewidth}
p{\dimexpr.10\linewidth-2\tabcolsep-1.3333\arrayrulewidth}
p{\dimexpr.10\linewidth-2\tabcolsep-1.3333\arrayrulewidth}
p{\dimexpr.10\linewidth-2\tabcolsep-1.3333\arrayrulewidth}
p{\dimexpr.10\linewidth-2\tabcolsep-1.3333\arrayrulewidth}
p{\dimexpr.10\linewidth-2\tabcolsep-1.3333\arrayrulewidth}
}
\toprule
 Classifier definition   & Exclude incorrect from ROC   & ROC truth       & Multi-class score   &   AUROC &   TPR @1\%FPR &   \% ID-incorrect* rejected &   \% ID-incorrect* rejected - min & OOD vs ID-incorrect* accuracy   & OOD vs ID-incorrect* F1 \\
\midrule
 Multi-class              & no                           & ID vs OOD       & ID                 &    98.8 &        67.4 &                        5.6 &                              5.6 & -                               & -                         \\\midrule
 ID vs OOD               & no                           & ID vs OOD       & ID                 &    98.7 &        68   &                        6   &                              6   & -                               & -                         \\\midrule
 Multi-class              & no                           & ID vs OOD       & ID-correct         &    98.4 &        70.6 &                        5.4 &                              4.9 & 98.5                            & 53.7                      \\\midrule
 ID-correct vs rest      & no                           & ID vs OOD       & ID                 &    98.3 &        70.6 &                        6   &                              6   & -                               & -                         \\\midrule
 ID-correct vs rest      & no                           & not(ID-correct) & ID                 &    98.4 &        83.6 &                       24.2 &                             24.2 & -                               & -                         \\\midrule
 Multi-class              & no                           & not(ID-correct) & ID                 &    97.5 &        83.7 &                       20.7 &                             20.7 & -                               & -                         \\\midrule
 ID vs OOD               & no                           & not(ID-correct) & ID                 &    97.3 &        84   &                       20.7 &                             20.7 & -                               & -                         \\\midrule
 Multi-class              & no                           & not(ID-correct) & ID-correct         &    98.4 &        84.1 &                       24.1 &                             21.5 & 96.2                            & 54.4                      \\\midrule
 ID-correct vs rest      & yes                          & not(ID-correct) & ID                 &    99.5 &        92.6 &                       24.2 &                             24.2 & -                               & -                         \\\midrule
 ID-correct vs rest      & yes                          & ID vs OOD       & ID                 &    99.5 &        92.6 &                       24.2 &                             24.2 & -                               & -                         \\\midrule
 Multi-class              & yes                          & not(ID-correct) & ID-correct         &    99.5 &        93.2 &                       24.1 &                             21.5 & 96.2                            & 54.4                      \\\midrule
 Multi-class              & yes                          & ID vs OOD       & ID-correct         &    99.5 &        93.2 &                       24.1 &                             21.5 & 96.2                            & 54.4                      \\\midrule
 Multi-class              & yes                          & not(ID-correct) & ID                 &    99.6 &        93.3 &                       20.7 &                             20.7 & -                               & -                         \\\midrule
 Multi-class              & yes                          & ID vs OOD       & ID                 &    99.6 &        93.3 &                       20.7 &                             20.7 & -                               & -                         \\\midrule
 ID vs OOD               & yes                          & not(ID-correct) & ID                 &    99.6 &        93.6 &                       20.7 &                             20.7 & -                               & -                         \\\midrule
 ID vs OOD               & yes                          & ID vs OOD       & ID                 &    99.6 &        93.6 &                       20.7 &                             20.7 & -                               & -                         \\
\bottomrule
\end{tabular}
    \caption{Effect of reference: MSM Norwegian vertebrates}
    \label{tab:my_label}
\end{table*}
\begin{table*}
    \centering
    \begin{tabular}{
p{\dimexpr.12\linewidth-2\tabcolsep-1.3333\arrayrulewidth}
p{\dimexpr.09\linewidth-2\tabcolsep-1.3333\arrayrulewidth}
p{\dimexpr.11\linewidth-2\tabcolsep-1.3333\arrayrulewidth}
p{\dimexpr.09\linewidth-2\tabcolsep-1.3333\arrayrulewidth}
p{\dimexpr.09\linewidth-2\tabcolsep-1.3333\arrayrulewidth}
p{\dimexpr.10\linewidth-2\tabcolsep-1.3333\arrayrulewidth}
p{\dimexpr.10\linewidth-2\tabcolsep-1.3333\arrayrulewidth}
p{\dimexpr.10\linewidth-2\tabcolsep-1.3333\arrayrulewidth}
p{\dimexpr.10\linewidth-2\tabcolsep-1.3333\arrayrulewidth}
p{\dimexpr.10\linewidth-2\tabcolsep-1.3333\arrayrulewidth}
}
\hline
 Classifier definition   & Exclude incorrect from ROC   & ROC truth       & Multi-class score   &   AUROC &   TPR @1\%FPR &   \% ID-incorrect* rejected &   \% ID-incorrect* rejected - min & OOD vs ID-incorrect* accuracy   & OOD vs ID-incorrect* F1   \\\midrule
\hline
 ID-correct vs rest      & no                           & ID vs OOD       & ID                 &    96.3 &        58.5 &                        0   &                              0   & -                               & -                         \\\midrule
 Multi-class              & no                           & ID vs OOD       & ID-correct         &    96.4 &        58.6 &                        0   &                              0   & -                               & -                         \\\midrule
 ID-correct vs rest      & no                           & not(ID-correct) & ID                 &    95.5 &        66.7 &                       10.7 &                             10.7 & -                               & -                         \\\midrule
 ID vs OOD               & no                           & not(ID-correct) & ID                 &    92.3 &        67.3 &                        7.8 &                              7.8 & -                               & -                         \\\midrule
 Multi-class              & no                           & not(ID-correct) & ID                 &    92.9 &        68.2 &                        8.8 &                              8.8 & -                               & -                         \\\midrule
 ID vs OOD               & no                           & ID vs OOD       & ID                 &    98   &        68.9 &                        2   &                              2   & -                               & -                         \\\midrule
 Multi-class              & no                           & not(ID-correct) & ID-correct         &    95.5 &        69.1 &                       12.7 &                              8.9 & 95.9                            & 64.1                      \\\midrule
 Multi-class              & no                           & ID vs OOD       & ID                 &    98   &        70.3 &                        2.2 &                              2.2 & -                               & -                         \\\midrule
 ID-correct vs rest      & yes                          & not(ID-correct) & ID                 &    98.9 &        85.4 &                       10.7 &                             10.7 & -                               & -                         \\\midrule
 ID-correct vs rest      & yes                          & ID vs OOD       & ID                 &    98.9 &        85.4 &                       10.7 &                             10.7 & -                               & -                         \\\midrule
 ID vs OOD               & yes                          & not(ID-correct) & ID                 &    99.2 &        87.7 &                        7.8 &                              7.8 & -                               & -                         \\\midrule
 ID vs OOD               & yes                          & ID vs OOD       & ID                 &    99.2 &        87.7 &                        7.8 &                              7.8 & -                               & -                         \\\midrule
 Multi-class              & yes                          & not(ID-correct) & ID-correct         &    98.8 &        87.9 &                       12.7 &                              8.9 & 95.9                            & 64.1                      \\\midrule
 Multi-class              & yes                          & ID vs OOD       & ID-correct         &    98.8 &        87.9 &                       12.7 &                              8.9 & 95.9                            & 64.1                      \\\midrule
 Multi-class              & yes                          & not(ID-correct) & ID                 &    99.2 &        88.5 &                        8.8 &                              8.8 & -                               & -                         \\\midrule
 Multi-class              & yes                          & ID vs OOD       & ID                 &    99.2 &        88.5 &                        8.8 &                              8.8 & -                               & -                         \\\midrule
\hline
\end{tabular}
    \caption{Effect of reference: : iNaturalist 2018}
    \label{tab:inat_cat_ood_app}
\end{table*}

\clearpage

\subsection{SHAP analysis}
\label{sec:shap_app}

\begin{minipage}{\textwidth}
    \captionsetup{type=figure}
    \centering
    \begin{subfigure}{0.48\linewidth}
        \includegraphics[width=1.0\textwidth]{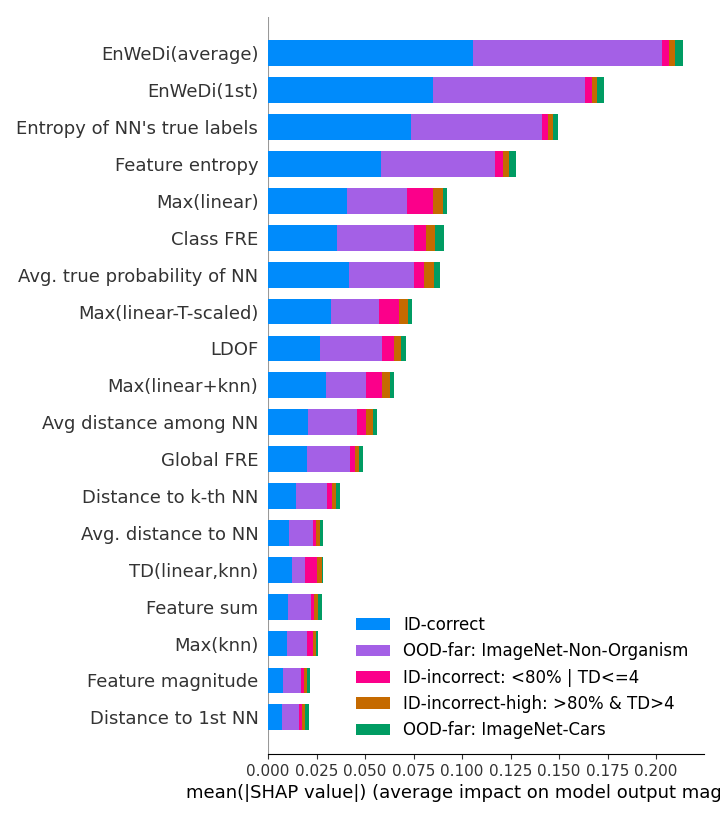}
        \caption{MSM top-level}
        \label{fig:shap_msm_top_app}
    \end{subfigure}
    \begin{subfigure}{0.48\linewidth}
        \includegraphics[width=1.0\textwidth]{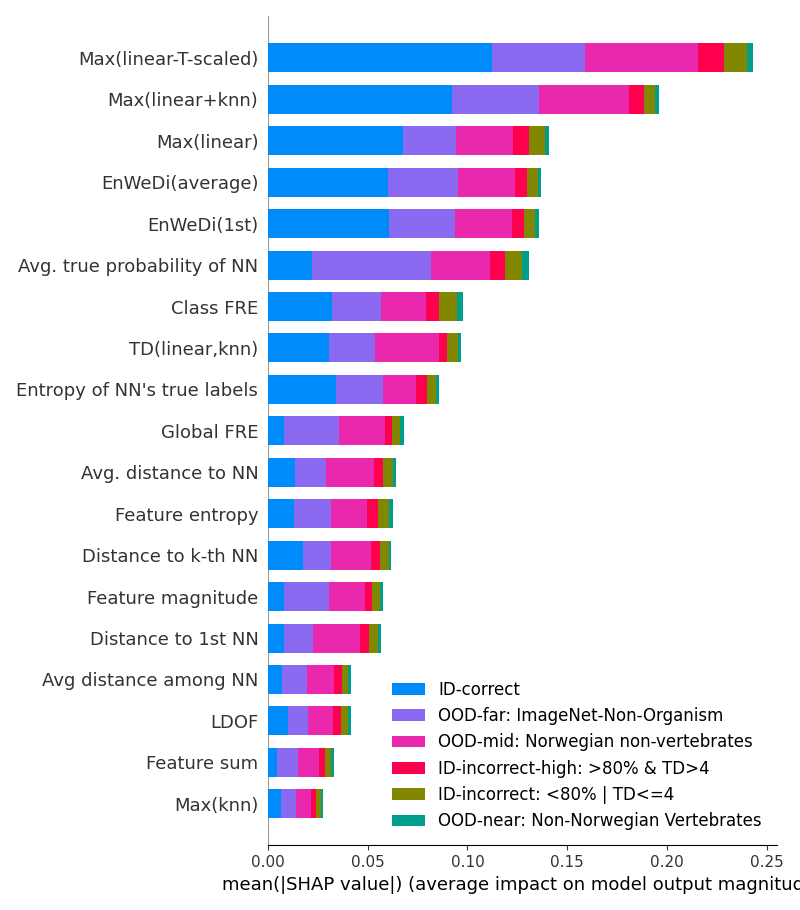}
        \caption{Norwegian vertebrates}
        \label{fig:shap_nor}
    \end{subfigure}
    \begin{subfigure}{0.50\linewidth}
        \includegraphics[width=1.0\textwidth]{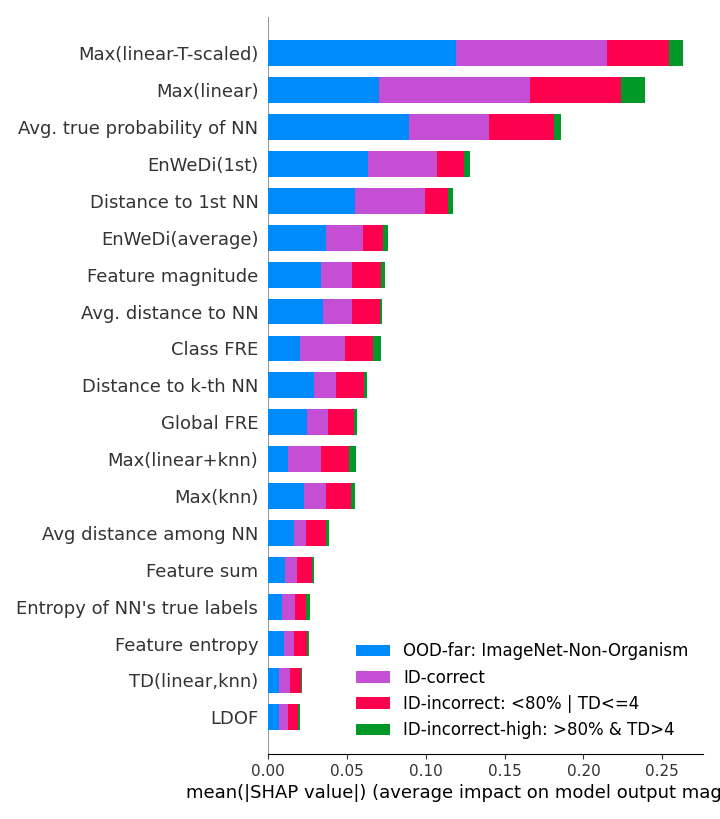}
        \caption{iNaturalist 2018}
        \label{fig:inat_shap_class}
    \end{subfigure}
    \caption{SHAP analysis showing individual OOD measures most contributing to the COOD model}
    \label{fig:shap_inat}
    \hfill
\end{minipage}

\clearpage

\subsubsection{SHAP analysis per output category}
\label{sec:shap_class_app}
\begin{minipage}{\textwidth}
    \captionsetup{type=figure}
    \centering
    \begin{subfigure}{0.45\linewidth}
        \includegraphics[width=1.0\textwidth]{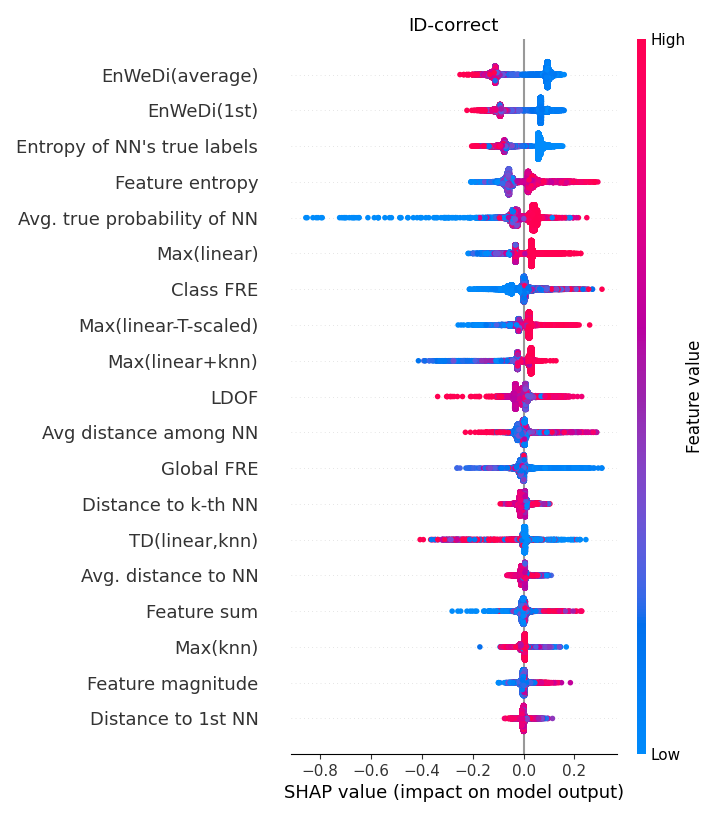}
    \end{subfigure}
    \hfill
    \begin{subfigure}{0.45\linewidth}
        \includegraphics[width=1.0\textwidth]{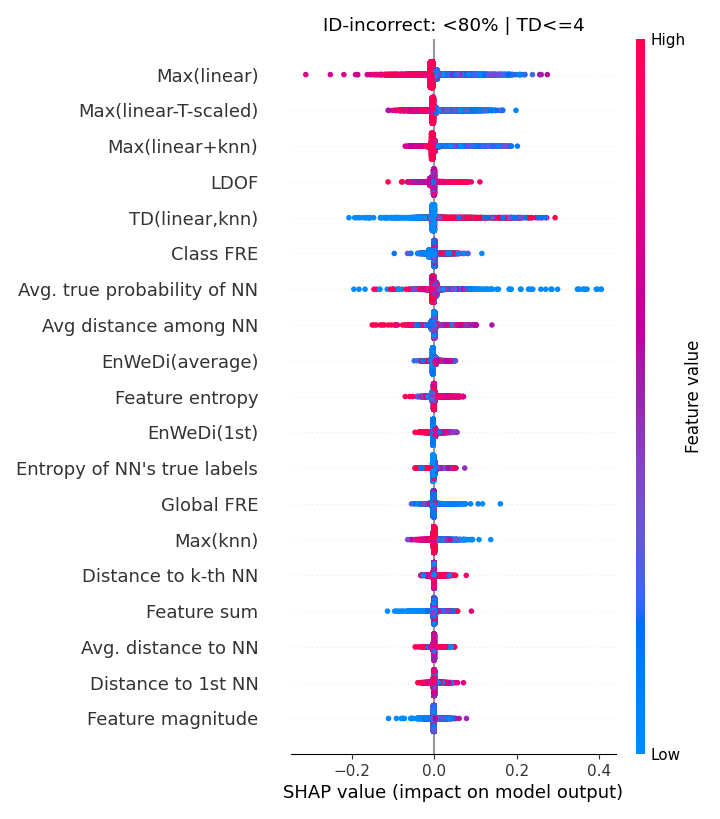}
    \end{subfigure}
    \begin{subfigure}{0.45\linewidth}
        \includegraphics[width=1.0\textwidth]{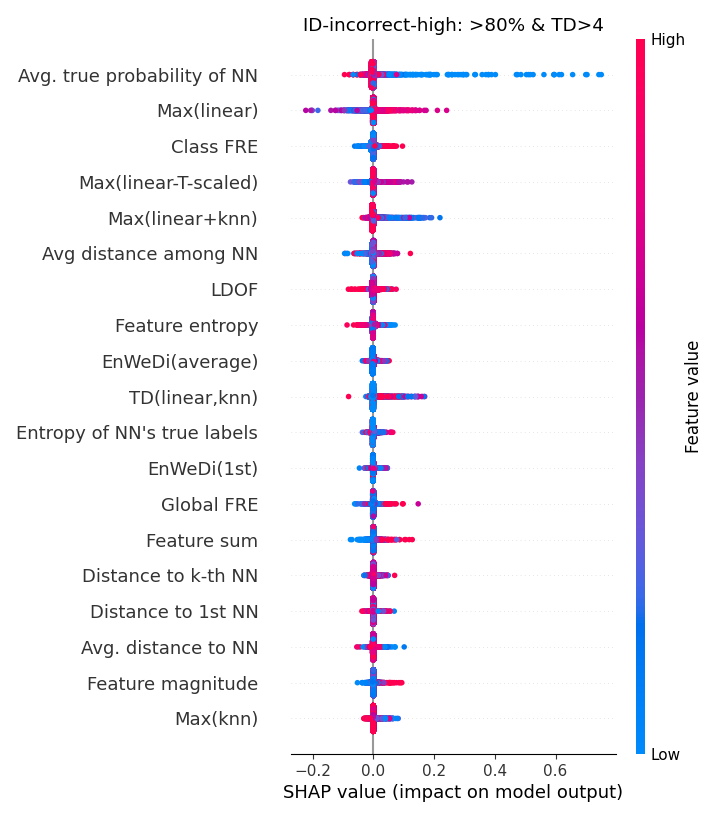}
    \end{subfigure}
    \hfill
    \begin{subfigure}{0.45\linewidth}
        \includegraphics[width=1.0\textwidth]{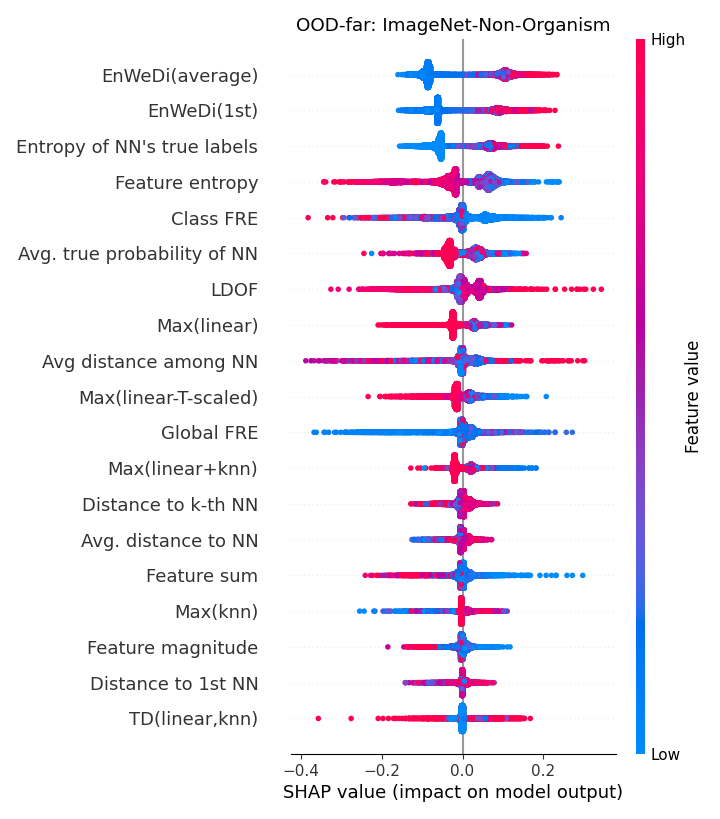}
    \end{subfigure}
    \caption{SHAP analysis per output category: MSM top-level}
\end{minipage}

\begin{figure*}
    \ContinuedFloat
    \centering
    \begin{subfigure}{0.45\linewidth}
        \includegraphics[width=1.0\textwidth]{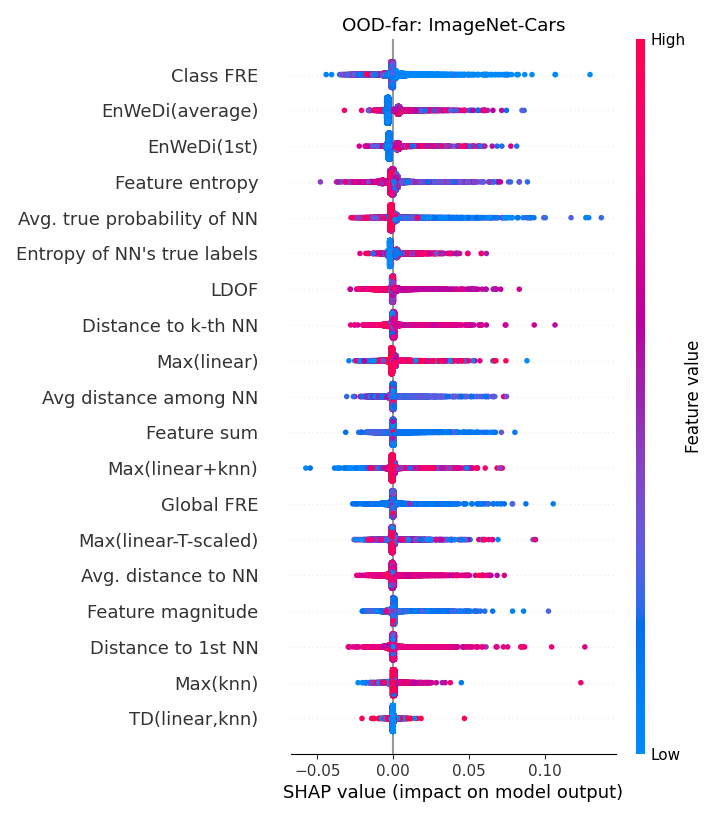}
    \end{subfigure}    
    \caption{SHAP analysis per output category: MSM top-level (cont.)}
    \label{tab:shap_top_app}
\end{figure*}

\begin{figure*}
    \centering
    \begin{subfigure}{0.48\linewidth}
        \includegraphics[width=1.0\textwidth]{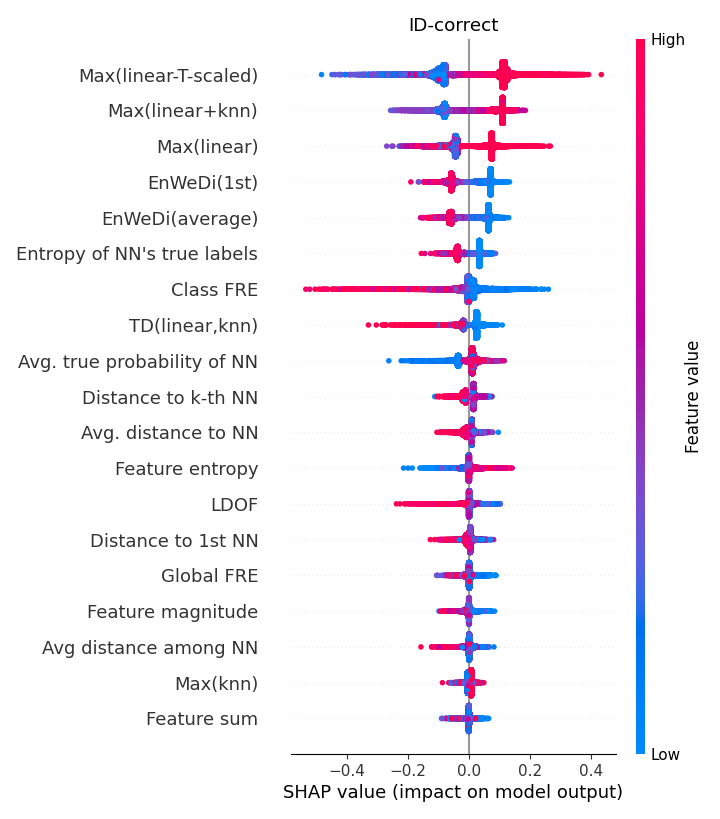}
    \end{subfigure}
    \begin{subfigure}{0.48\linewidth}
        \includegraphics[width=1.0\textwidth]{chordata_adb_k30/shap_id-correct.png}
    \end{subfigure}
    \begin{subfigure}{0.48\linewidth}
        \includegraphics[width=1.0\textwidth]{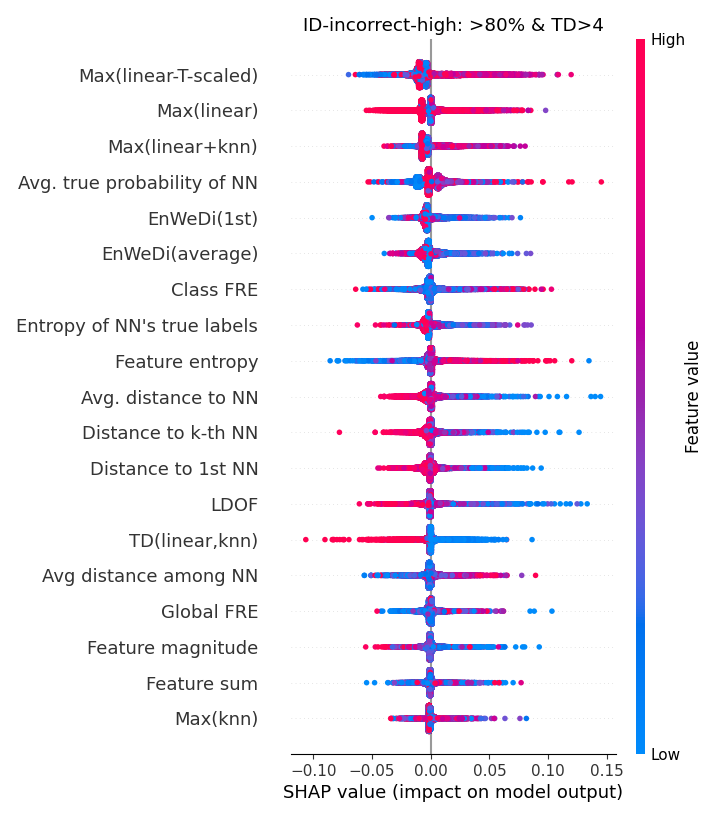}
    \end{subfigure}
    \begin{subfigure}{0.48\linewidth}
        \includegraphics[width=1.0\textwidth]{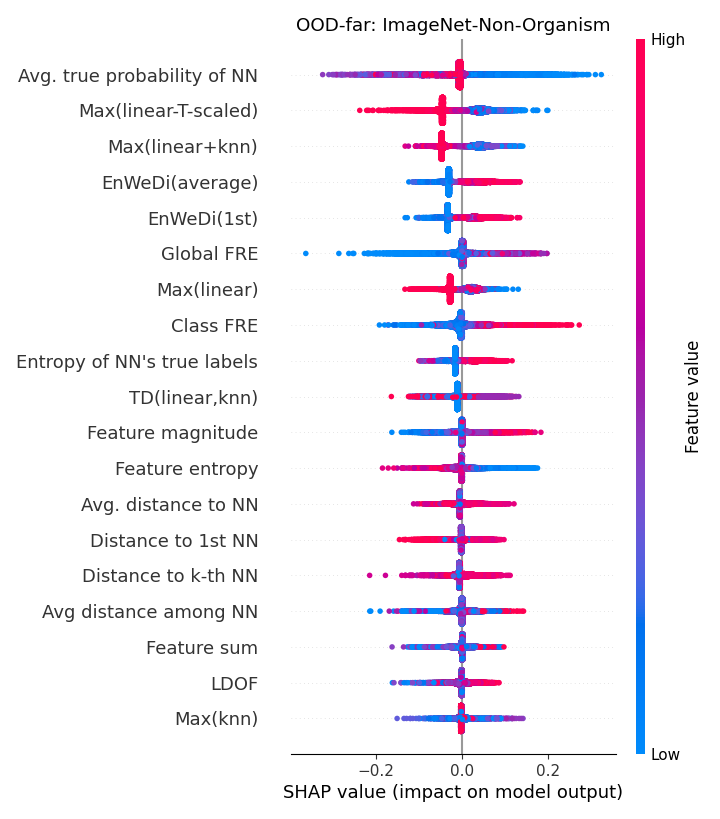}
    \end{subfigure}
    
    \caption{SHAP analysis per output category: Norwegian vertebrates}
\end{figure*}

\begin{figure*}
    \ContinuedFloat
    \centering
    \begin{subfigure}{0.48\linewidth}
        \includegraphics[width=1.0\textwidth]{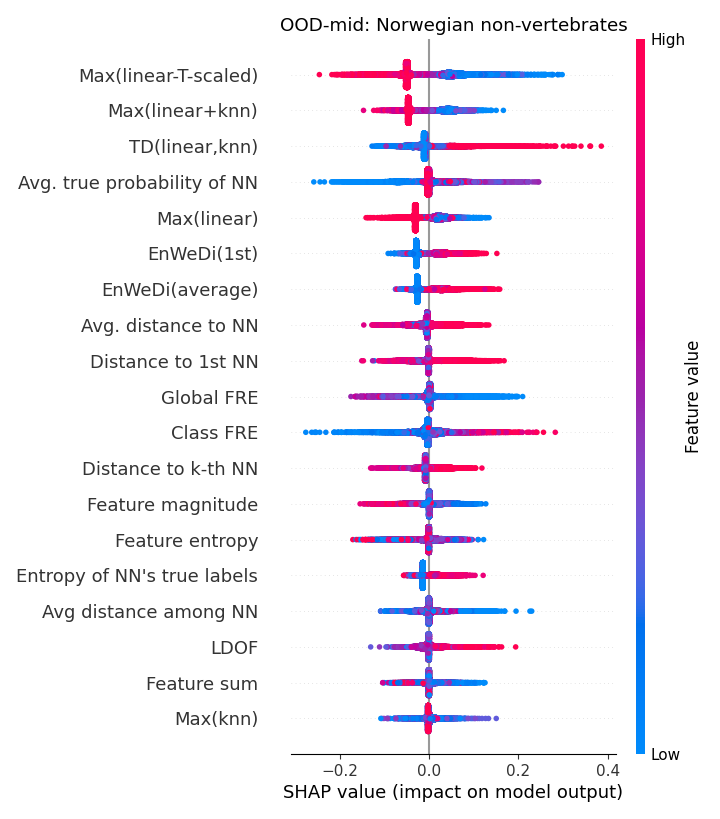}
    \end{subfigure}
    \begin{subfigure}{0.48\linewidth}
        \includegraphics[width=1.0\textwidth]{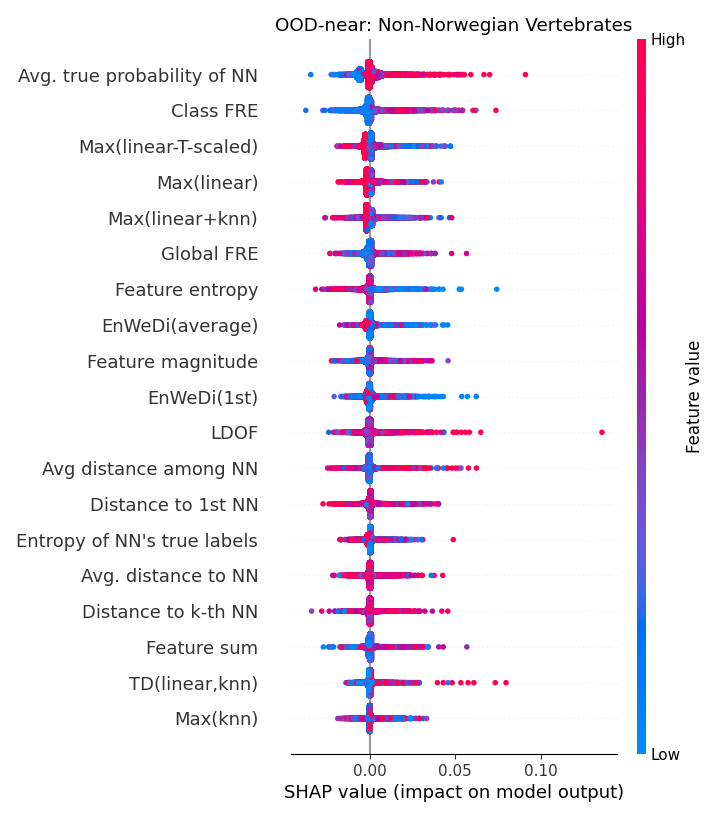}
    \end{subfigure}
    \caption{SHAP analysis per output category: Norwegian vertebrates (cont.)}
\end{figure*}

\begin{figure*}
    \centering
    \begin{subfigure}{0.45\linewidth}
        \includegraphics[width=1.0\textwidth]{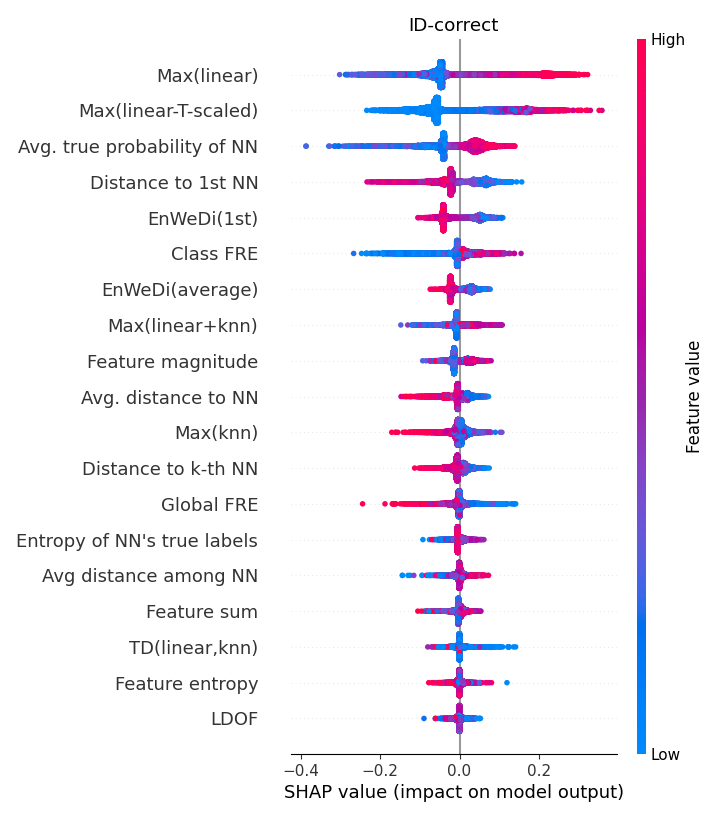}
    \end{subfigure}
    \hfill
    \begin{subfigure}{0.45\linewidth}
        \includegraphics[width=1.0\textwidth]{inat__k30/shap_id-correct.png}
    \end{subfigure}
    \begin{subfigure}{0.45\linewidth}
        \includegraphics[width=1.0\textwidth]{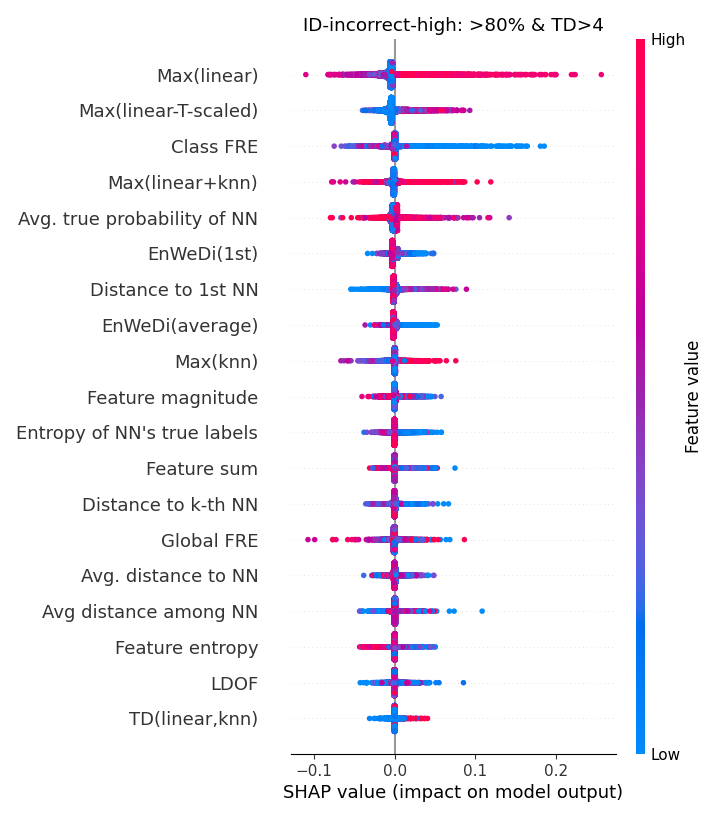}
    \end{subfigure}
    \hfill
    \begin{subfigure}{0.45\linewidth}
        \includegraphics[width=1.0\textwidth]{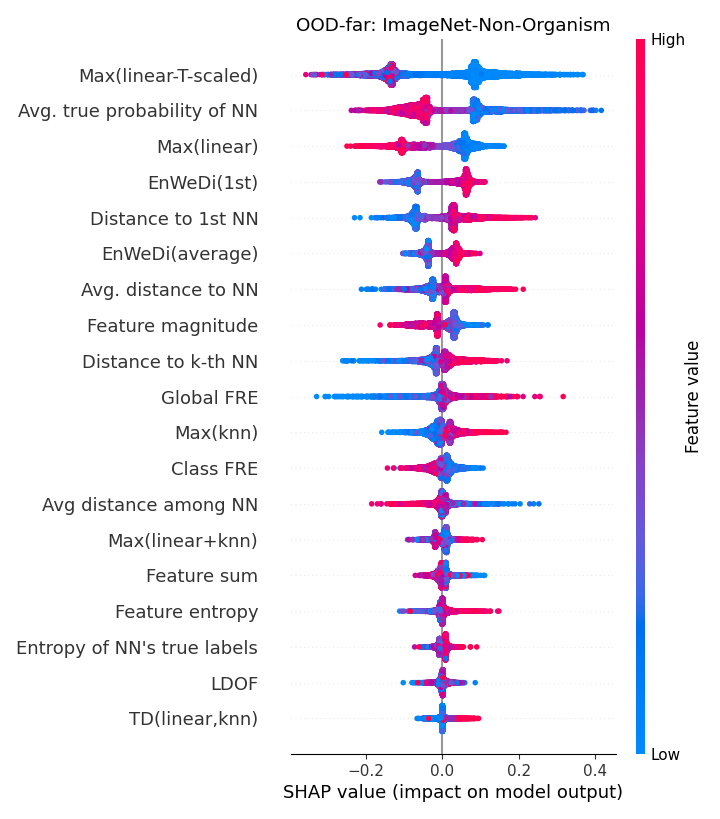}
    \end{subfigure}
    \caption{SHAP analysis per output category: iNaturalist 2018}
\end{figure*}

\clearpage
\onecolumn
\subsection{COOD score distribution and ROC analysis}



\begin{figure*}[hbt!]  
    \includegraphics[width=1.00\textwidth]{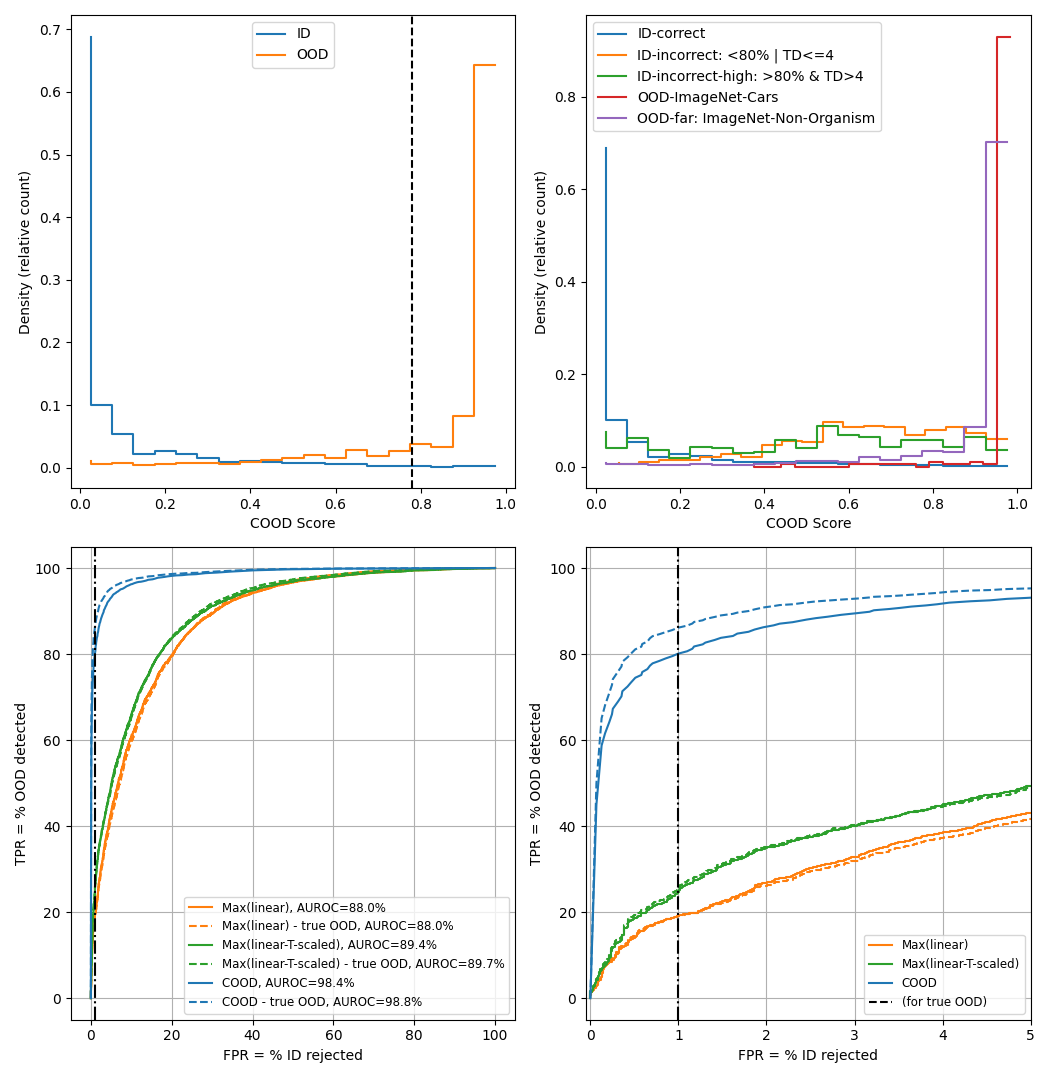}
    \caption{COOD score distribution and ROC analysis for MSM top-level}
\end{figure*}

\begin{figure*}[hbt!]
    \includegraphics[width=1.00\textwidth]{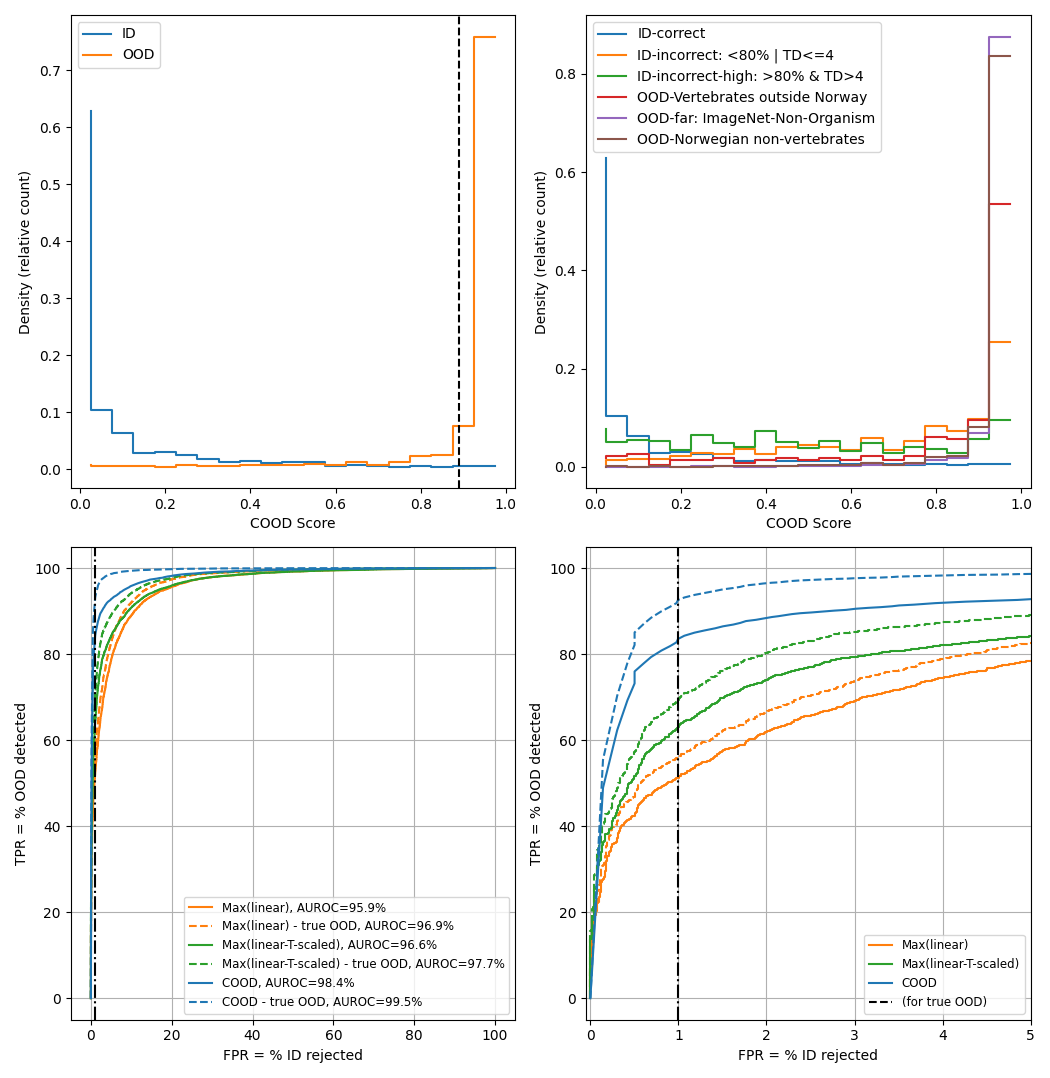}
    \caption{COOD score distribution and ROC analysis for Norwegian vertebrates}
\end{figure*}

\begin{figure*}[hbt!]
    \includegraphics[width=1.00\textwidth]{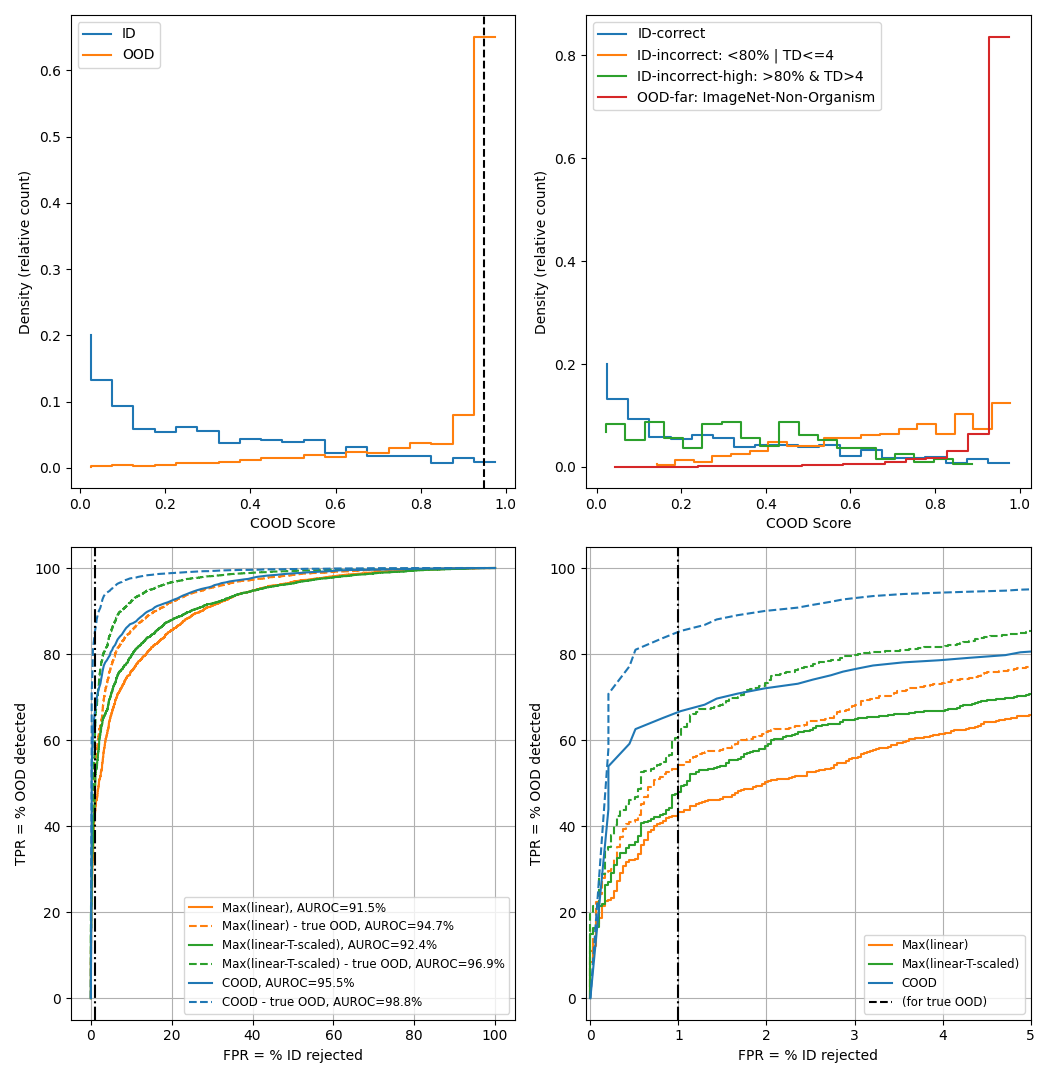}
    \caption{COOD score distribution and ROC analysis for iNaturalist 2018}
\end{figure*}

\clearpage

\label{app:example_images}
\subsection{Example images}

\vspace{2cm}

\begin{minipage}{\textwidth}
    \captionsetup{type=figure}
    \centering
    \begin{subfigure}{\linewidth}
        \centering
        \includegraphics[width=1.0\textwidth]{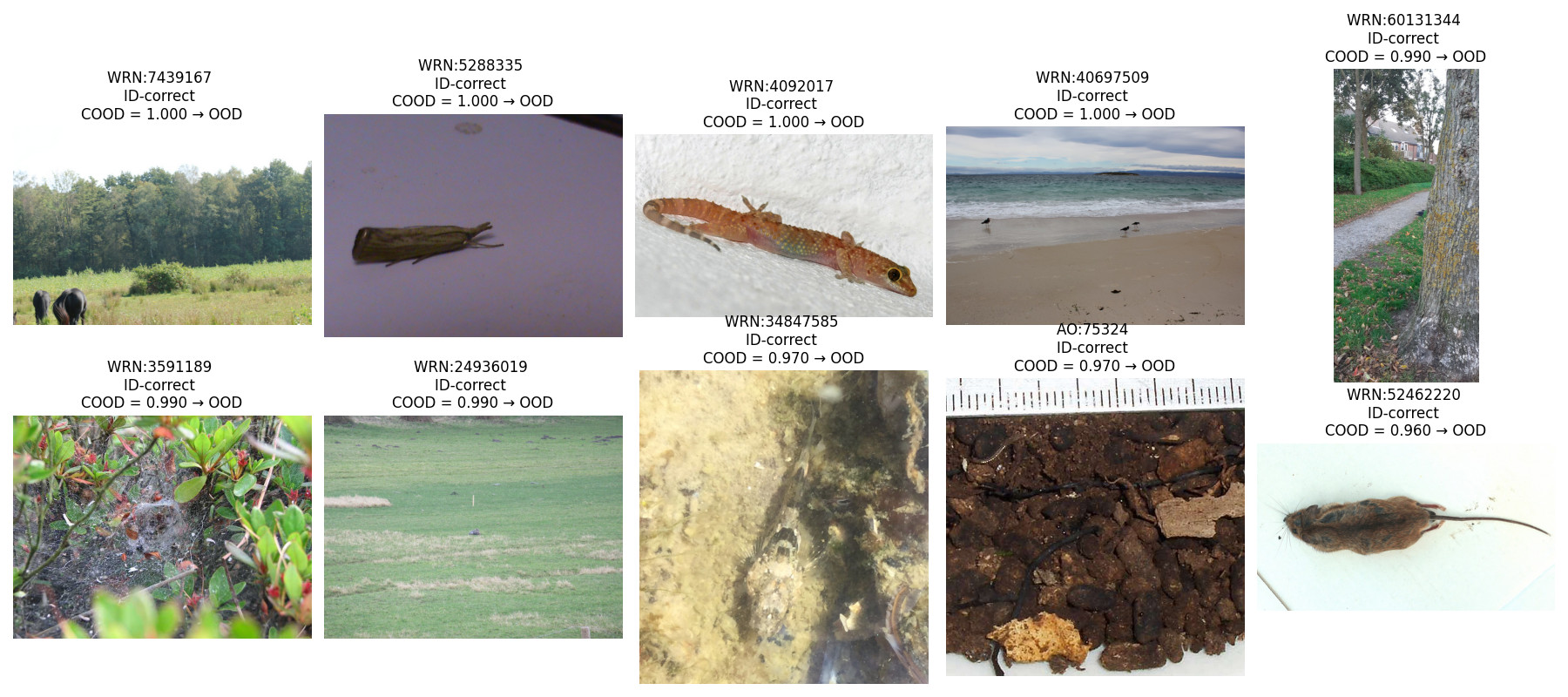}
        \caption{False positives (high COOD score for ID images)}
        
    \end{subfigure}
    \hfill
    \begin{subfigure}{\linewidth}
        \centering
        \includegraphics[width=1.0\textwidth]{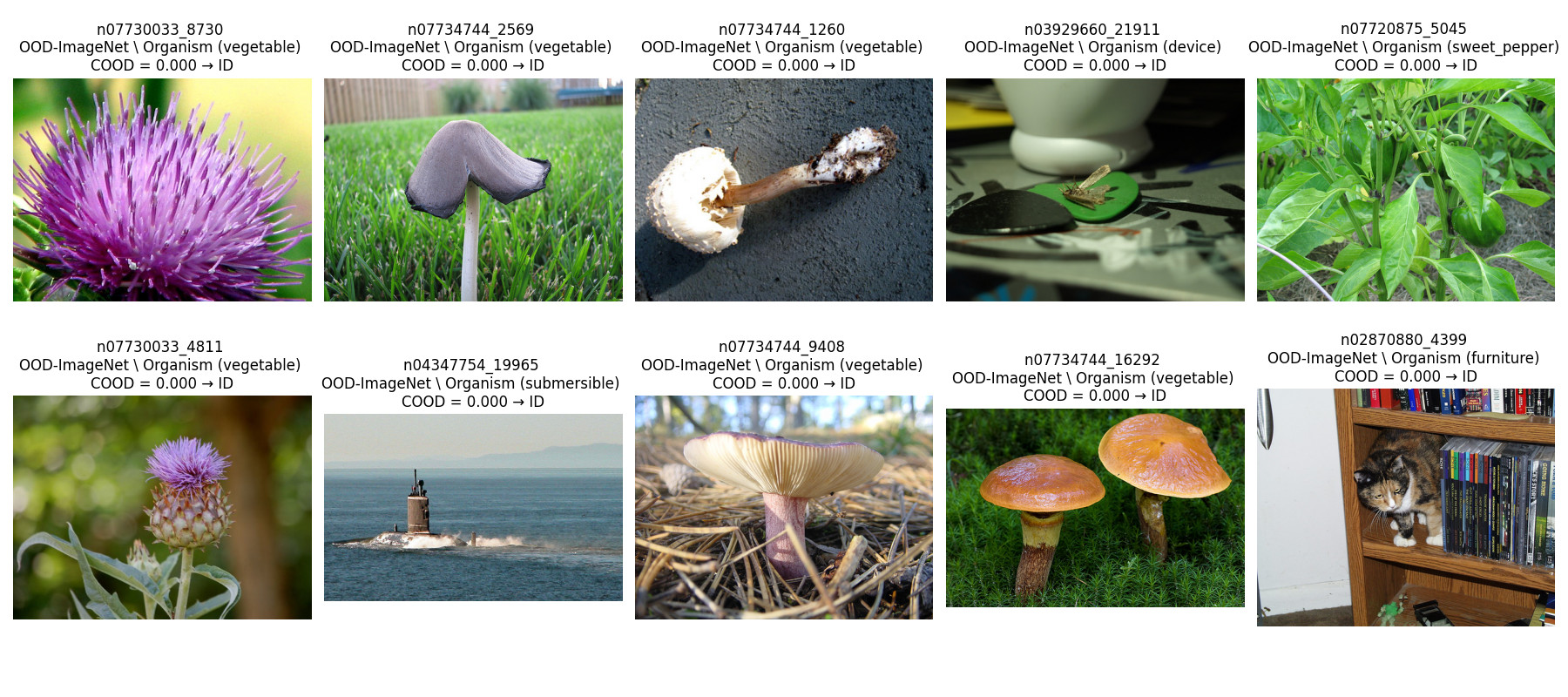}
        \caption{False negatives (low COOD score for OOD images)}
        
    \end{subfigure}
    \caption{Selected examples of MSM top-level}
    \label{fig:msm-fp-fn}
\end{minipage}

\begin{figure*}
    \centering
    \begin{subfigure}{\linewidth}
        \centering
        \includegraphics[width=1.0\textwidth]{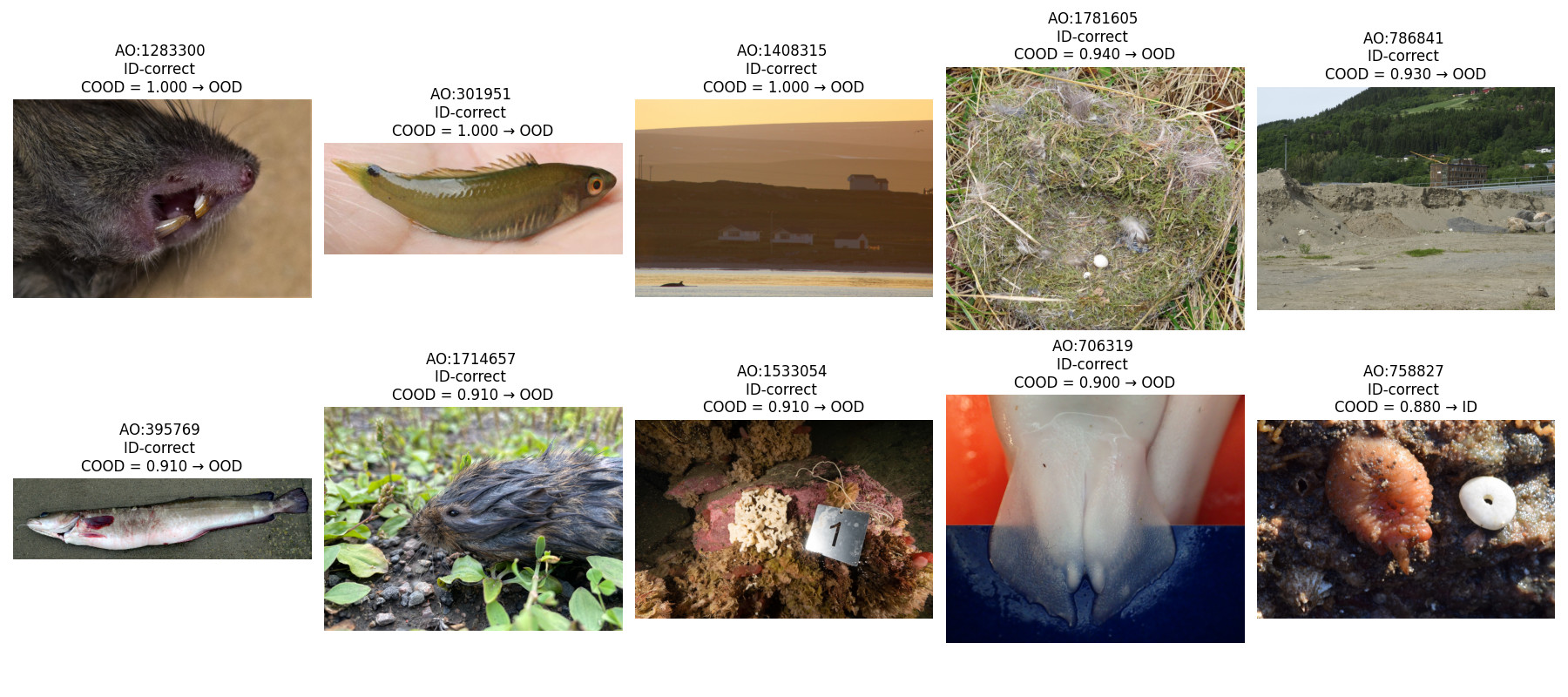}
        \caption{False positives (high COOD score for ID images)}
    \end{subfigure}
    \hfill
    \begin{subfigure}{\linewidth}
        \centering
        \includegraphics[width=1.0\textwidth]{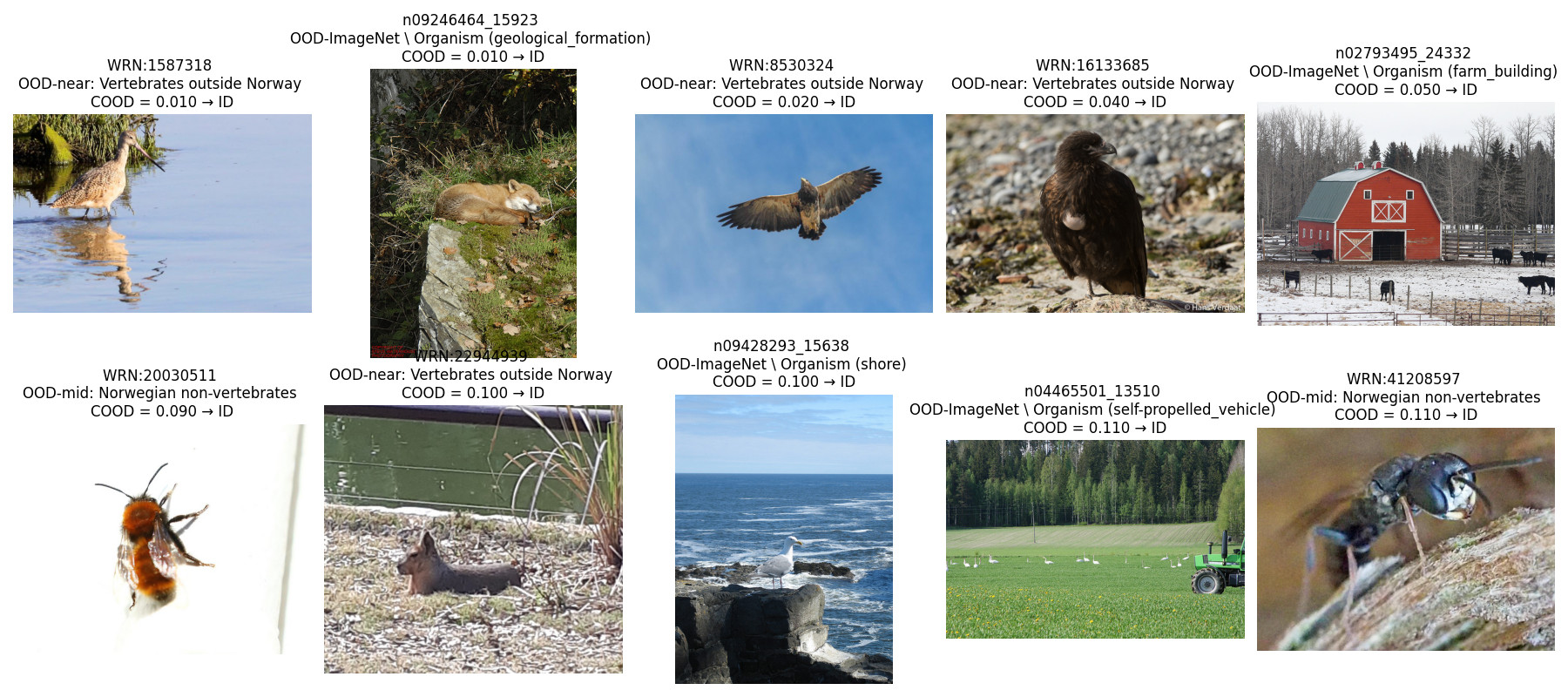}
        \caption{False negatives (low COOD score for OOD images)}
    \end{subfigure}
    \caption{Selected examples of Norwegian vertebrates}
    \label{fig:norwegian-fp-fn}  
\end{figure*}

\clearpage

\subsection{Visual example licences}

\vspace{3cm}
\begin{minipage}{\textwidth}
    \centering
    \captionsetup{type=table}
    \begin{tabular}{lllll}
\toprule
 uid          & author              & date       & license     & image\_url                                    \\
\midrule
 WRN:7439167  & Ted van der Knaap   & 2014-09-18 & CC BY-NC-ND & waarneming.nl/photos/7439167         \\
 WRN:60131344 & Jona Haasnoot       & 2022-10-24 & CC BY-NC-ND & waarneming.nl/photos/60131344        \\
 WRN:24936019 & Theo Ruppert        & 2020-02-20 & CC BY-NC-ND & waarneming.nl/photos/24936019        \\
 WRN:40697509 & KJ Hijlkema         & 2018-03-12 & CC BY-NC-ND & waarneming.nl/photos/40697509        \\
 WRN:4092017  & Harm Alberts        & 2012-10-01 & CC BY-NC-ND & waarneming.nl/photos/4092017         \\
 WRN:3591189  & Veerle De Saedeleer & 2012-07-07 & CC BY-NC-ND & waarneming.nl/photos/3591189         \\
 WRN:5288335  & Ruben Vernieuwe     & 2013-08-09 & CC BY-NC-ND & waarneming.nl/photos/5288335         \\
 WRN:34847585 & Jeroen Hoek         & 2021-04-28 & CC BY-NC-ND & waarneming.nl/photos/34847585        \\
 AO:75324     & Magne Flåten        & 2008-06-17 & CC BY-SA    & artsobservasjoner.no/Image/75324 \\
 WRN:52462220 & Cor de Jong         & 2022-06-20 & CC BY-NC-ND & waarneming.nl/photos/52462220        \\
\bottomrule
\end{tabular}
    \caption{MSM top-level: License information for visual examples}
    \label{tab:used_examples_msm_top}
\end{minipage}
\hfill
\vspace{2cm}

\begin{minipage}{\textwidth}
    \centering
    \captionsetup{type=table}
    \begin{tabular}{lllll}
\toprule
 uid          & author                & date       & license     & image\_url                                      \\
\midrule
 AO:301951    & Magne Flåten          & 2012-02-09 & CC BY-SA    & artsobservasjoner.no/Image/301951  \\
 AO:1408315   & Karel Samyn           & 2020-07-13 & CC BY-NC-SA & artsobservasjoner.no/Image/1408315 \\
 WRN:1587318  & Frans Rosmalen        & 2010-09-24 & CC BY-NC-ND & waarneming.nl/photos/1587318           \\
 WRN:8530324  & Paul Schrijvershof    & 2015-03-22 & CC BY-NC    & waarneming.nl/photos/8530324           \\
 WRN:41208597 & Jos Cuppens           & 2021-08-14 & CC BY-NC-ND & waarneming.nl/photos/41208597          \\
 WRN:22944939 & David Tempelman       & 2019-08-25 & CC0         & waarneming.nl/photos/22944939          \\
 WRN:20030511 & Frens Westenbrink     & 2019-03-31 & CC BY-NC-ND & waarneming.nl/photos/20030511          \\
 AO:1283300   & Johan Sirnes          & 2020-01-17 & CC BY       & artsobservasjoner.no/Image/1283300 \\
 WRN:16133685 & Hans Verdaat          & 2018-01-21 & CC BY-NC-ND & waarneming.nl/photos/16133685          \\
 AO:1714657   & Eric Francois Roualet & 2021-06-08 & CC BY-NC-SA & artsobservasjoner.no/Image/1714657 \\
 AO:395769    & Kristoffer Bøhn       & 2014-01-03 & CC BY-NC-SA & artsobservasjoner.no/Image/395769  \\
 AO:1781605   & Magne Flåten          & 2013-12-09 & CC BY-SA    & artsobservasjoner.no/Image/1781605 \\
 AO:1533054   & Ole Meldahl           & 2020-12-26 & CC BY-SA    & artsobservasjoner.no/Image/1533054 \\
 AO:706319    & Magne Flåten          & 2016-11-29 & CC BY-SA    & artsobservasjoner.no/Image/706319  \\
 AO:786841    & Anders Breili         & 2017-06-17 & CC BY-NC-SA & artsobservasjoner.no/Image/786841  \\
 AO:758827    & Johan Sirnes          & 2017-04-29 & CC BY       & artsobservasjoner.no/Image/758827  \\
\bottomrule
\end{tabular}
    \caption{Norwegian vertebrates: License information for visual examples}
    \label{tab:used_examples_adb}
\end{minipage}

\end{document}